\documentclass{article}
\pdfoutput=1
\usepackage{times}
\usepackage{graphicx} 
\usepackage{subfigure}
\usepackage{natbib}
\usepackage{amsmath}
\usepackage{enumerate}
\usepackage{bbm}
\usepackage{makecell}
\usepackage{booktabs}
\usepackage{tabularx}
\usepackage{ulem}
\usepackage[table]{xcolor}
\usepackage{savesym}
\savesymbol{TrimSpaces}
\usepackage{siunitx}
\restoresymbol{SI}{TrimSpaces}
\usepackage{csvsimple}
\usepackage{longtable}
\usepackage{multicol}
\usepackage{chngcntr}

\renewcommand{\cite}{\citep}

\usepackage{bibunits}
\defaultbibliography{icml2015-arxiv}
\defaultbibliographystyle{icml2015}

\DeclareMathOperator*{\Mean}{Mean}
\DeclareMathOperator*{\Median}{Median}

\usepackage{algorithm}
\usepackage{algorithmic}
\usepackage{hyperref}

\usepackage[accepted]{icml2015}

\icmltitlerunning{Massively Multitask Networks for Drug Discovery}
\title{Massively Multitask Networks for Drug Discovery: Appendix}

\begin{document}
\begin{bibunit}
\twocolumn[
\icmltitle{Massively Multitask Networks for Drug Discovery}

\icmlauthor{Bharath Ramsundar\textsuperscript{*,$\dagger$, $\circ$}}{rbharath@stanford.edu}
\icmlauthor{Steven Kearnes\textsuperscript{*,$\dagger$}}{kearnes@stanford.edu}
\icmlauthor{Patrick Riley\textsuperscript{$\circ$}}{pfr@google.com}
\icmlauthor{Dale Webster\textsuperscript{$\circ$}}{drw@google.com}
\icmlauthor{David Konerding\textsuperscript{$\circ$}}{dek@google.com}
\icmlauthor{Vijay Pande\textsuperscript{$\dagger$}}{pande@stanford.edu}
\icmladdress{(\textsuperscript{*}Equal
contribution, \textsuperscript{$\dagger$}Stanford
University, \textsuperscript{$\circ$}Google Inc.)}

\icmlkeywords{deep learning, bioinformatics}

\vskip 0.3in
]

\begin{abstract}
Massively multitask neural architectures provide a learning framework for
drug discovery that synthesizes information from many distinct biological
sources. To train these architectures at scale, we gather large amounts of
data from public sources to create a dataset of nearly 40 million
measurements across more than 200 biological targets. We investigate
several aspects of the multitask framework by performing a series of
empirical studies and obtain some interesting results: $(1)$ massively
multitask networks obtain predictive accuracies significantly better than
single-task methods, $(2)$ the predictive power of multitask networks
improves as additional tasks and data are added, $(3)$ the total amount of
data and the total number of tasks both contribute significantly to
multitask improvement, and $(4)$ multitask networks afford limited
transferability to tasks not in the training set. Our results underscore
the need for greater data sharing and further algorithmic innovation to
accelerate the drug discovery process.
\end{abstract}

\section{Introduction}
\label{intro}

Discovering new treatments for human diseases is an immensely complicated
challenge. Prospective drugs must attack the source of an illness, but must
do so while satisfying restrictive metabolic and toxicity constraints.
Traditionally, drug discovery is an extended process that takes years to
move from start to finish, with high rates of failure along the way.

After a suitable target has been identified, the first step in the drug
discovery process is ``hit finding.'' Given some druggable target,
pharmaceutical companies will screen millions of drug-like compounds in an
effort to find a few attractive molecules for further optimization. These
screens are often automated via robots, but are expensive to perform.
Virtual screening attempts to replace or augment the high-throughput
screening process by the use of computational methods
\cite{shoichet2004virtual}. Machine learning methods have frequently been
applied to virtual screening by training supervised classifiers to predict
interactions between targets and small molecules.

There are a variety of challenges that must be overcome to achieve
effective virtual screening. Low hit rates in experimental screens (often
only $1$--$2$\% of screened compounds are active against a given target)
result in imbalanced datasets that require special handling for effective
learning. For instance, care must be taken to guard against unrealistic
divisions between active and inactive compounds (``artificial enrichment'')
and against information leakage due to strong similarity between active
compounds (``analog bias'') \cite{rohrer2009maximum}.  Furthermore, the
paucity of experimental data means that overfitting is a perennial thorn.

The overall complexity of the virtual screening problem has limited the
impact of machine learning in drug discovery. To achieve greater predictive
power, learning algorithms must combine disparate sources of experimental
data across multiple targets. Deep learning provides a flexible paradigm
for synthesizing large amounts of data into predictive models. In
particular, multitask networks facilitate information sharing across
different experiments and compensate for the limited data associated with
any particular experiment.

In this work, we investigate several aspects of the multitask learning
paradigm as applied to virtual screening. We gather a large collection of
datasets containing nearly 40 million experimental measurements for over
200 targets. We demonstrate that multitask networks trained on this
collection achieve significant improvements over baseline machine learning
methods. We show that adding more tasks and more data yields better
performance. This effect diminishes as more data and tasks are added, but
does not appear to plateau within our collection. Interestingly, we find
that the total amount of data and the total number of tasks both have
significant roles in this improvement. Furthermore, the features extracted
by the multitask networks demonstrate some transferability to tasks not
contained in the training set. Finally, we find that the presence of shared
active compounds is moderately correlated with multitask improvement, but
the biological class of the target is not.

\section{Related Works}

Machine learning has a rich history in drug discovery. Early work combined
creative featurizations of molecules with off-the-shelf learning algorithms
to predict drug activity \cite{varnek2012machine}. The state of the art has
moved to more refined models, such as the influence relevance voting method
that combines low-complexity neural networks and k-nearest neighbors
\cite{swamidass2009influence}, and Bayesian belief networks that repurpose
textual information retrieval methods for virtual screening
\cite{abdo2010ligand}. Other related work uses deep recursive neural
networks to predict aqueous solubility by extracting features from the
connectivity graphs of small molecules \cite{lusci2013deep}.

Deep learning has made inroads into drug discovery in recent years, most
notably in 2012 with the Merck Kaggle competition \cite{dahl2012deep}.
Teams were given pre-computed molecular descriptors for compounds with
experimentally measured activity against 15 targets and were asked to
predict the activity of molecules in a held-out test set. The winning team
used ensemble models including multitask deep neural networks, Gaussian
process regression, and dropout to improve the baseline test set $R^2$ by
nearly 17\%. The winners of this contest later released a technical report
that discusses the use of multitask networks for virtual screening
\cite{dahl2014multi}. Additional work at Merck analyzed the choice of
hyperparameters when training single- and multitask networks and showed
improvement over random forest models \cite{ma2015deep}. The Merck Kaggle
result has been received with skepticism by some in the cheminformatics and
drug discovery communities \citep[and associated comments]{lowe2012did}.
Two major concerns raised were that the sample size was too small (a good
result across 15 systems may well have occurred by chance) and that any
gains in predictive accuracy were too small to justify the increase in
complexity.

While we were preparing this work, a workshop paper was released that also
used massively multitask networks for virtual screening
\cite{unterthinerdeep}. That work curated a dataset of 1,280 biological
targets with 2 million associated data points and trained a multitask
network. Their network has more tasks than ours (1,280 vs. 259) but far
fewer data points (2 million vs. nearly 40 million). The emphasis of our
work is considerably different; while their report highlights the
performance gains due to multitask networks, ours is focused on
disentangling the underlying causes of these improvements. Another closely
related work proposed the use of collaborative filtering for virtual
screening and employed both multitask networks and kernel-based methods
\cite{erhan2006collaborative}. Their multitask networks, however, did not
consistently outperform single-task models.

Within the greater context of deep learning, we draw upon various strands
of recent thought. Prior work has used multitask deep networks in the
contexts of language understanding \cite{collobert2008unified} and
multi-language speech recognition \cite{deng2013new}. Our best-performing
networks draw upon design patterns introduced by GoogLeNet
\cite{szegedy2014going}, the winner of ILSVRC 2014.

\section{Methods}

\subsection{Dataset Construction and Design}
\label{sec:datasets}
Models were trained on $259$ datasets gathered from publicly available
data.  These datasets were divided into four groups: PCBA, MUV, DUD-E, and
Tox21.  The PCBA group contained $128$ experiments in the PubChem BioAssay
database \cite{wang2012pubchem}.  The MUV group contained $17$ challenging
datasets specifically designed to avoid common pitfalls in virtual
screening \cite{rohrer2009maximum}. The DUD-E group contained $102$
datasets that were designed for the evaluation of methods to predict
interactions between proteins and small molecules
\cite{mysinger2012directory}.  The Tox21 datasets were used in the recent
Tox21 Data Challenge (\url{https://tripod.nih.gov/tox21/challenge/}) and
contained experimental data for $12$ targets relevant to drug toxicity
prediction. We used only the training data from this challenge because the
test set had not been released when we constructed our collection.  In
total, our $259$ datasets contained $37.8$M experimental data points for
$1.6$M compounds. Details for the dataset groups are given in
\tablename~\ref{tab:datasets}. See the Appendix for details on individual
datasets and their biological target categorization.

\begin{table}
\small
\centering
\caption{Details for dataset groups. Values for the number of data points
per dataset and the percentage of active compounds are reported as means,
with standard deviations in parenthesis.}
\vskip 0.2 in
\label{tab:datasets}
\begin{tabular}{llll}
\toprule
{Group} & {Datasets} & {Data Points / ea.} & {\% Active} \\
\midrule
{PCBA} & 128 & $282\text{K}$ $(122\text{K})$ & $1.8$ $(3.8)$ \\
DUD-E & $102$ & $14\text{K}$ $(11\text{K})$ & $1.6$ $(0.2)$ \\
MUV & $17$ & $15\text{K}$ $(1)$ & $0.2$ $(0)$ \\
Tox21 & $12$ & $6\text{K}$ $(500)$ & $7.8$ $(4.7)$ \\
\bottomrule
\end{tabular}
\vskip -0.2in
\end{table}

It should be noted that we did not perform any preprocessing of our
datasets, such as removing potential experimental artifacts. Such artifacts
may be due by compounds whose physical properties cause interference with
experimental measurements or allow for promiscuous interactions with many
targets. A notable exception is the MUV group, which has been processed
with consideration of these pathologies \cite{rohrer2009maximum}.

\subsection{Small Molecule Featurization}
We used extended connectivity fingerprints (ECFP4)
\cite{rogers2010extended} generated by RDKit \cite{landrumrdkit} to
featurize each molecule. The molecule is decomposed into a set of
fragments---each centered at a non-hydrogen atom---where each fragment
extends radially along bonds to neighboring atoms. Each fragment is
assigned a unique identifier, and the collection of identifiers for a
molecule is hashed into a fixed-length bit vector to construct the
molecular ``fingerprint''.  ECFP4 and other fingerprints are commonly used
in cheminformatics applications, especially to measure similarity between
compounds \cite{willett1998chemical}. A number of molecules (especially in
the Tox21 group) failed the featurization process and were not used in
training our networks. See the Appendix for details.

\subsection{Validation Scheme and Metrics}

The traditional approach for model evaluation is to have fixed training,
validation, and test sets.  However, the imbalance present in our datasets
means that performance varies widely depending on the particular
training/test split. To compensate for this variability, we used stratified
$K$-fold cross-validation; that is, each fold maintains the active/inactive
proportion present in the unsplit data. For the remainder of the paper, we
use $K=5$.

Note that we did not choose an explicit validation set. Several datasets in
our collection have very few actives ($\sim30$ each for the MUV group), and
we feared that selecting a specific validation set would skew our results.
As a consequence, we suspect that our choice of hyperparameters may be
affected by information leakage across folds. However, our networks do not
appear to be highly sensitive to hyperparameter choice (see
Section~\ref{sec:experimental}), so we do not consider leakage to be a
serious issue.

Following recommendations from the cheminformatics community
\cite{jain2008recommendations}, we used metrics derived from the receiver
operating characteristic (ROC) curve to evaluate model performance. Recall
that the ROC curve for a binary classifier is the plot of true positive
rate (TPR) vs. false positive rate (FPR) as the discrimination threshold is
varied.  For individual datasets, we are interested in the area under the
ROC curve (AUC), which is a global measure of classification performance
(note that AUC must lie in the range $[0, 1]$).  More generally, for a
collection of $N$ datasets, we consider the mean and median
$K$-fold-average AUC:
\[
\Mean/\Median\left \{\frac{1}{K} \sum_{k=1}^K \text{AUC}_k(D_n)\middle | \ n=1,\dotsc,N\right \},
\]
where $\text{AUC}_k(D_n)$ is defined as the AUC of a classifier trained on
folds $\{1,\dotsc,K\} \setminus k$ of dataset $D_n$ and tested on fold $k$.
For completeness, we include in the Appendix an alternative metric called
``enrichment'' that is widely used in the cheminformatics literature
\cite{jain2008recommendations}. We note that many other performance metrics
exist in the literature; the lack of standard metrics makes it difficult to
do direct comparisons with previous work.

\subsection{Multitask Networks}

\begin{figure}[ht]
\centering
\includegraphics[trim=0 4.5in 5.5in 0,clip,width=0.9\linewidth]{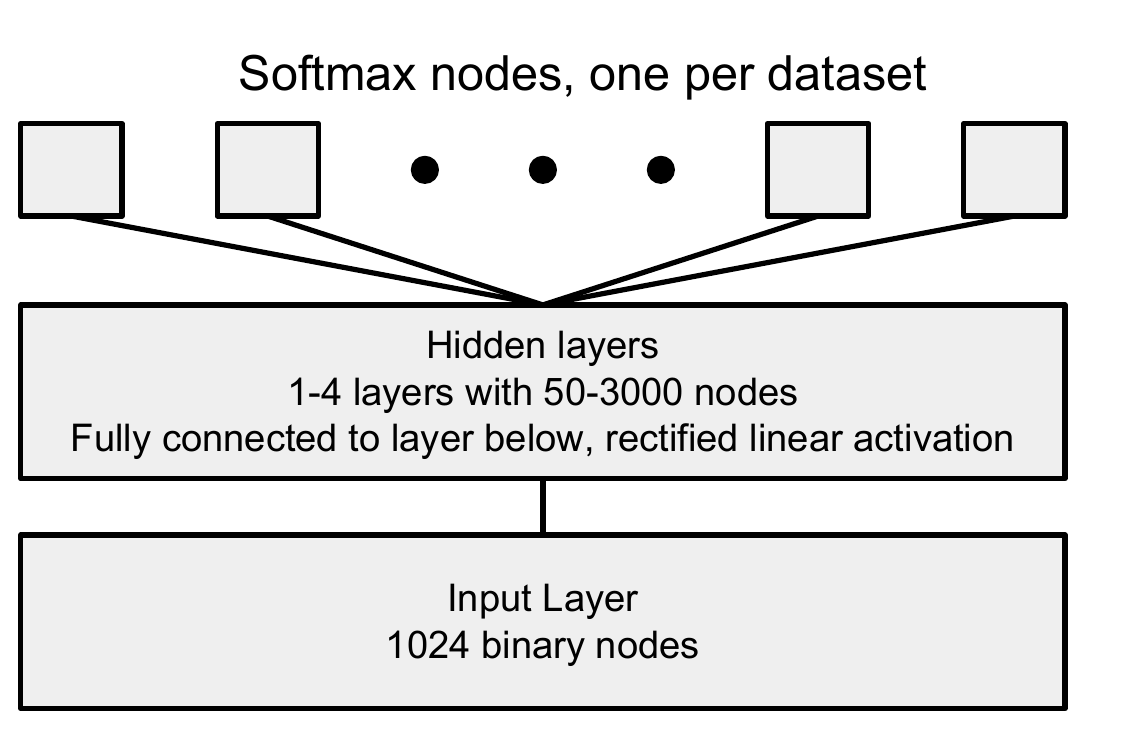}
\caption{Multitask neural network.}
\label{fig:network}
\end{figure}

A neural network is a nonlinear classifier that performs repeated linear
and nonlinear transformations on its input. Let $\mathbf{x}_i$ represent
the input to the $i$-th layer of the network (where $\mathbf{x}_0$ is
simply the feature vector). The transformation performed is
\[
\mathbf{x}_{i+1} = \sigma(\mathbf{W}_{i} \mathbf{x}_i + \mathbf{b}_{i})
\]
where $\mathbf{W}_i$ and $\mathbf{b}_i$ are respectively the weight matrix
and bias for the $i$-th layer, and $\sigma$ is a nonlinearity (in our work,
the rectified linear unit \cite{nair2010rectified}). After $L$ such
transformations, the final layer of the network $\mathbf{x}_L$ is then fed
to a simple linear classifier, such as the softmax, which predicts the
probability that the input $\mathbf{x}_0$ has label $j$:
\[
P(y = j | \mathbf{x}_0 ) = \frac{e^{(\mathbf{w}^j)^T \mathbf{x}_L}}{\sum_{m=1}^M e^{(\mathbf{w}^m)^T \mathbf{x}_L}},
\]
where $M$ is the number of possible labels (here $M = 2$) and
$\mathbf{w}^{1}, \cdots, \mathbf{w}^{M}$ are weight vectors.
$\mathbf{W}_i$, $\mathbf{b}_i$, and $\mathbf{w}^m$ are learned during
training by the backpropagation algorithm \cite{rumelhart1988learning}.  A
multitask network attaches $N$ softmax classifiers, one for each task, to
the final layer $\mathbf{x}_L$. (A ``task'' corresponds to the classifier
associated with a particular dataset in our collection, although we often
use ``task'' and ``dataset'' interchangeably. See
\figurename~\ref{fig:network}.)

\section{Experimental Section}

In this section, we seek to answer a number of questions about the
performance, capabilities, and limitations of massively multitask neural
networks:

\begin{enumerate}
\itemsep0em
\item Do massively multitask networks provide a performance boost over
  simple machine learning methods? If so, what is the optimal architecture
  for massively multitask networks?
\item How does the performance of a multitask network depend on the number
  of tasks? How does the performance depend on the total amount of data?
\item Do massively multitask networks extract generalizable information
  about chemical space?
\item When do datasets benefit from multitask training?
\end{enumerate}

The following subsections detail a series of experiments that seek to
answer these questions.

\subsection{Experimental Exploration of Massively Multitask Networks}
\label{sec:experimental}
We investigate the performance of multitask networks with various
hyperparameters and compare to several standard machine learning
approaches. Table~\ref{tab:exp_results} shows some of the highlights of our
experiments. Our best multitask architecture (pyramidal multitask networks)
significantly outperformed simpler models, including a hypothetical model
whose performance on each dataset matches that of the best single-task
model (Max\{LR, RF, STNN, PSTNN\}).

Every model we trained performed extremely well on the DUD-E datasets (all
models in Table~\ref{tab:exp_results} had median $5$-fold-average AUCs
$\ge0.99$), making comparisons between models on DUD-E uninformative. For
that reason, we exclude DUD-E from our subsequent statistical analysis.
However, we did not remove DUD-E from the training altogether because doing
so adversely affected performance on the other datasets (data not shown);
we theorize that DUD-E helped to regularize the classifier and avoid
overfitting.

\begin{table*}[t]
\small
\caption{Median $5$-fold-average AUCs for various models.  For each model,
  the sign test in the last column estimates the fraction of datasets
  (excluding the DUD-E group, for reasons discussed in the text) for which
  that model is superior to the PMTNN (bottom row). We use the Wilson score interval to
  derive a $95\%$ confidence interval for this fraction.  Non-neural
  network methods were trained using scikit-learn
  \cite{pedregosa2011scikit} implementations and basic hyperparameter
  optimization. We also include results for a hypothetical ``best''
  single-task model (Max\{LR, RF, STNN, PSTNN\}) to provide a stronger
  baseline. Details for our cross-validation and training procedures are
  given in the Appendix.}
\label{tab:exp_results}
\vskip 0.2 in
\centering
\begin{tabular}{lcccc}
\toprule
Model & \makecell{PCBA \\ $(n=128)$} & \makecell{MUV \\ $(n=17)$} &
\makecell{Tox21 \\ $(n=12)$} & \makecell{Sign Test \\ CI} \\
\midrule
Logistic Regression (LR)& $.801$  & $.752$ & $.738$ & $[.04, .13]$ \\
Random Forest (RF) & $.800$  & $.774$  & $.790$ & $[.06, .16]$ \\
Single-Task Neural Net (STNN) & $.795$ & $.732$ & $.714$ & $[.04, .12]$ \\
Pyramidal $(2000, 100)$ STNN (PSTNN) & .809 & .745 & .740 & $[.06, .16]$ \\
Max\{LR, RF, STNN, PSTNN\} & $.824$ & $.781$ & $.790$ & $[.12, .24]$ \\
$1$-Hidden $(1200)$ Layer Multitask Neural Net (MTNN) & $.842$ & $.797$ & $.785$ & $[.08, .18]$ \\
Pyramidal $(2000, 100)$ Multitask Neural Net (PMTNN) & $\mathbf{.873}$ &
$\mathbf{.841}$ & $\mathbf{.818}$ & \\
\bottomrule
\end{tabular}
\end{table*}

During our first explorations, we had consistent problems with the networks
overfitting the data. As discussed in Section~\ref{sec:datasets}, our
datasets had a very small fraction of positive examples. For the single
hidden layer multitask network in Table~\ref{tab:exp_results}, each dataset
had $1200$ associated parameters. With a total number of positives in the
tens or hundreds, overfitting this number of parameters is a major issue in
the absence of strong regularization.

Reducing the number of parameters specific to each dataset is the
motivation for the pyramidal architecture. In our pyramidal networks, the
first hidden layer is very wide ($2000$ nodes) with a second narrow hidden
layer ($100$ nodes). This dimensionality reduction is similar in motivation
and implementation to the $1$x$1$ convolutions in the GoogLeNet
architecture~\cite{szegedy2014going}. The wide lower layer allows for
complex, expressive features to be learned while the narrow layer limits
the parameters specific to each task. Adding dropout of $0.25$ to our
pyramidal networks improved performance. We also trained single-task
versions of our best pyramidal network to understand whether this design
pattern works well with less data. Table~\ref{tab:exp_results} indicates
that these models outperform vanilla single-task networks but do not
substitute for multitask training.  Results for a variety of alternate
models are presented in the Appendix.

We investigated the sensitivity of our results to the sizes of the
pyramidal layers by running networks with all combinations of hidden layer
sizes: $(1000, 2000, 3000)$ and $(50, 100, 150)$.  Across the
architectures, means and medians shifted by $\le.01$~AUC with only MUV
showing larger changes with a range of $.038$.  We note that performance is
sensitive to the choice of learning rate and the number of training steps.
See the Appendix for details and data.

\subsection{Relationship between performance and number of tasks}
\label{sec:growth_curve}

\begin{figure}[ht]
\vskip 0.2in
\begin{center}
\centerline{\includegraphics[width=0.7\linewidth]{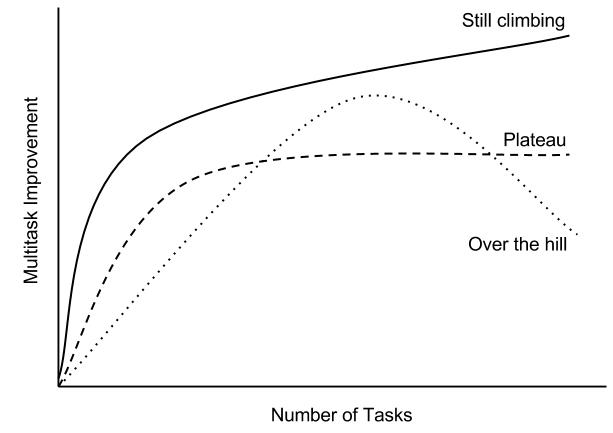}}
\caption{Potential multitask growth curves}
\label{fig:growth_held_in}
\end{center}
\vskip -0.2in
\end{figure}

The previous section demonstrated that massively multitask networks improve
performance over single-task models. In this section, we seek to understand
how multitask performance is affected by increasing the number of tasks.
\emph{A priori}, there are three reasonable ``growth curves'' (visually
represented in Figure~\ref{fig:growth_held_in}):
\begin{description}
\itemsep0em
\item[Over the hill:] performance initially improves, hits a maximum, then falls.
\item[Plateau:] performance initially improves, then plateaus.
\item[Still climbing:] performance improves throughout, but with a diminishing rate of
return.
\end{description}

We constructed and trained a series of multitask networks on datasets
containing $10, 20, 40, 80, 160,$ and $249$ tasks. These datasets all
contain a fixed set of ten ``held-in'' tasks, which consists of a randomly
sampled collection of five PCBA, three MUV, and two Tox21 datasets.  These
datasets correspond to unique targets that do not have any obvious analogs
in the remaining collection. (We also excluded a similarly chosen set of
ten ``held-out'' tasks for use in Section~\ref{sec:embedding}). Each
training collection is a superset of the preceding collection, with tasks
added randomly. For each network in the series, we computed the mean
$5$-fold-average-AUC for the tasks in the held-in collection. We repeated
this experiment ten times with different choices of random seed.

\begin{figure}[ht]
\centering
\includegraphics[width=\linewidth]{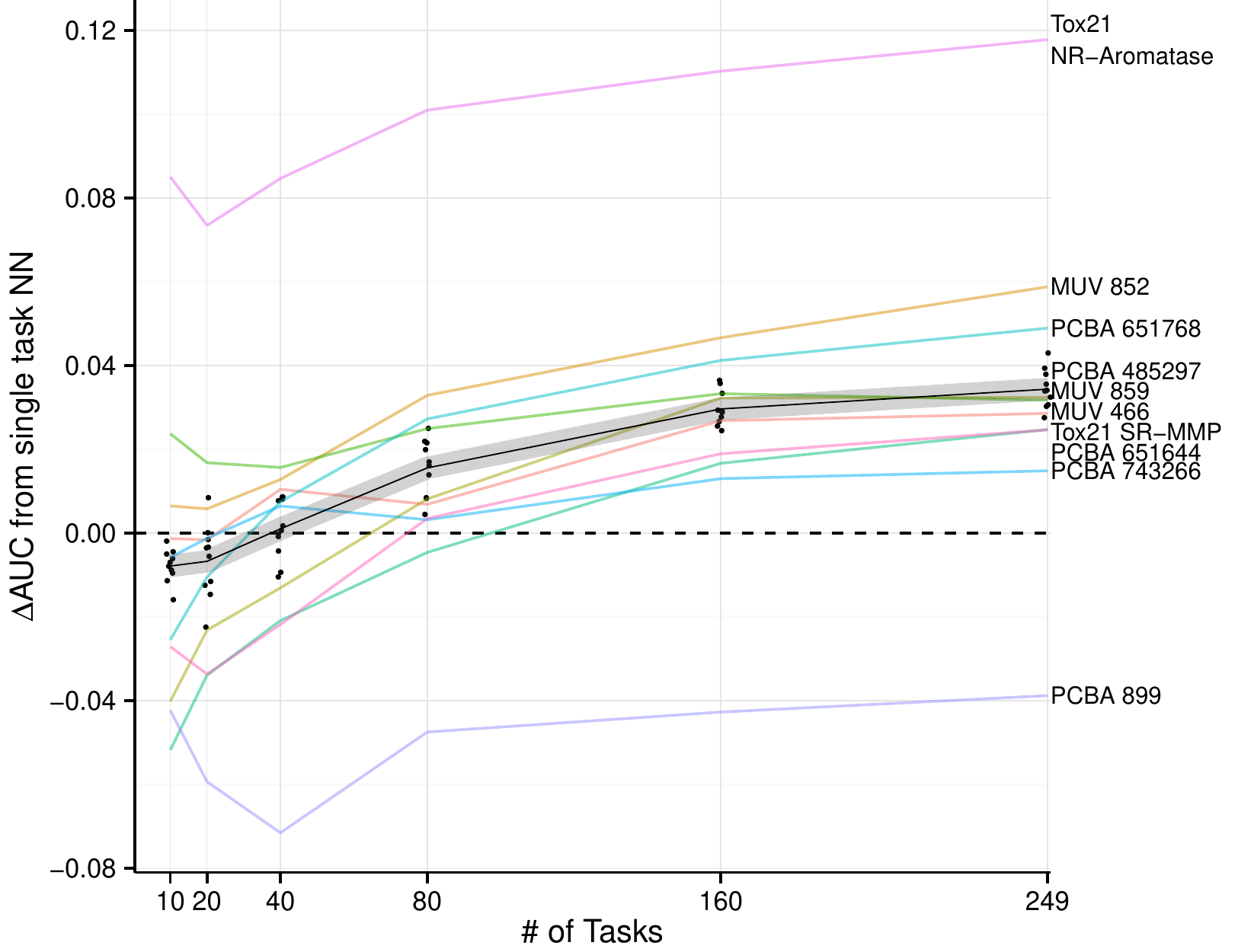}
\caption{Held-in growth curves. The $y$-axis shows the change in AUC
  compared to a single-task neural network with the same architecture
  (PSTNN).  Each colored curve is the multitask improvement for a given
  held-in dataset. Black dots represent means across the $10$ held-in
  datasets for each experimental run, where additional tasks were randomly
  selected.  The shaded curve is the mean across the $100$ combinations of
  datasets and experimental runs.}
\label{fig:held-in}
\end{figure}

Figure~\ref{fig:held-in} plots the results of our experiments. The shaded
region emphasizes the average growth curve, while black dots indicate
average results for different experimental runs. The figure also displays
lines associated with each held-in dataset. Note that several datasets show
initial dips in performance. However, all datasets show subsequent
improvement, and all but one achieves performance superior to the
single-task baseline. Within the limits of our current dataset collection,
the distribution in \figurename~\ref{fig:held-in} agrees with either
plateau or still climbing. The mean performance on the held-in set is still
increasing at $249$ tasks, so we hypothesize that performance is
\textbf{still climbing}. It is possible that our collection is too small
and that an alternate pattern may eventually emerge.

\subsection{More tasks or more data?}

In the previous section we studied the effects of adding more tasks, but
here we investigate the relative importance of the total amount of data vs.
the total number of tasks. Namely, is it better to have many tasks with a
small amount of associated data, or a small number of tasks with a large
amount of associated data?

We constructed a series of multitask networks with $10, 15, 20, 30, 50$ and
$82$ tasks. As in the previous section, the tasks are randomly associated
with the networks in a cumulative manner (\emph{i.e.}, the $82$-task
network contained all tasks present in the $50$-task network, and so on).
All networks contained the ten held-in tasks described in the previous
section. The $82$ tasks chosen were associated with the largest datasets in
our collection, each containing $300$K-$500$K data points. Note that all of
these tasks belonged to the PCBA group.

We then trained this series of networks multiple times with $1.6$M, $3.3$M,
$6.5$M, $13$M, and $23$M data points sampled from the non-held-in tasks. We
perform the sampling such that for a given task, all data points present in
the first stage ($1.6$M) appeared in the second ($3.3$M), all data points
present in the second stage appeared in the third ($6.5$M), and so on. We
decided to use larger datasets so we could sample meaningfully across this
entire range. Some combinations of tasks and data points were not realized;
for instance, we did not have enough data to train a $20$-task network with
$23$M additional data points. We repeated this experiment ten times using
different random seeds.

\begin{figure}[ht]
\centering
\includegraphics[width=\linewidth]{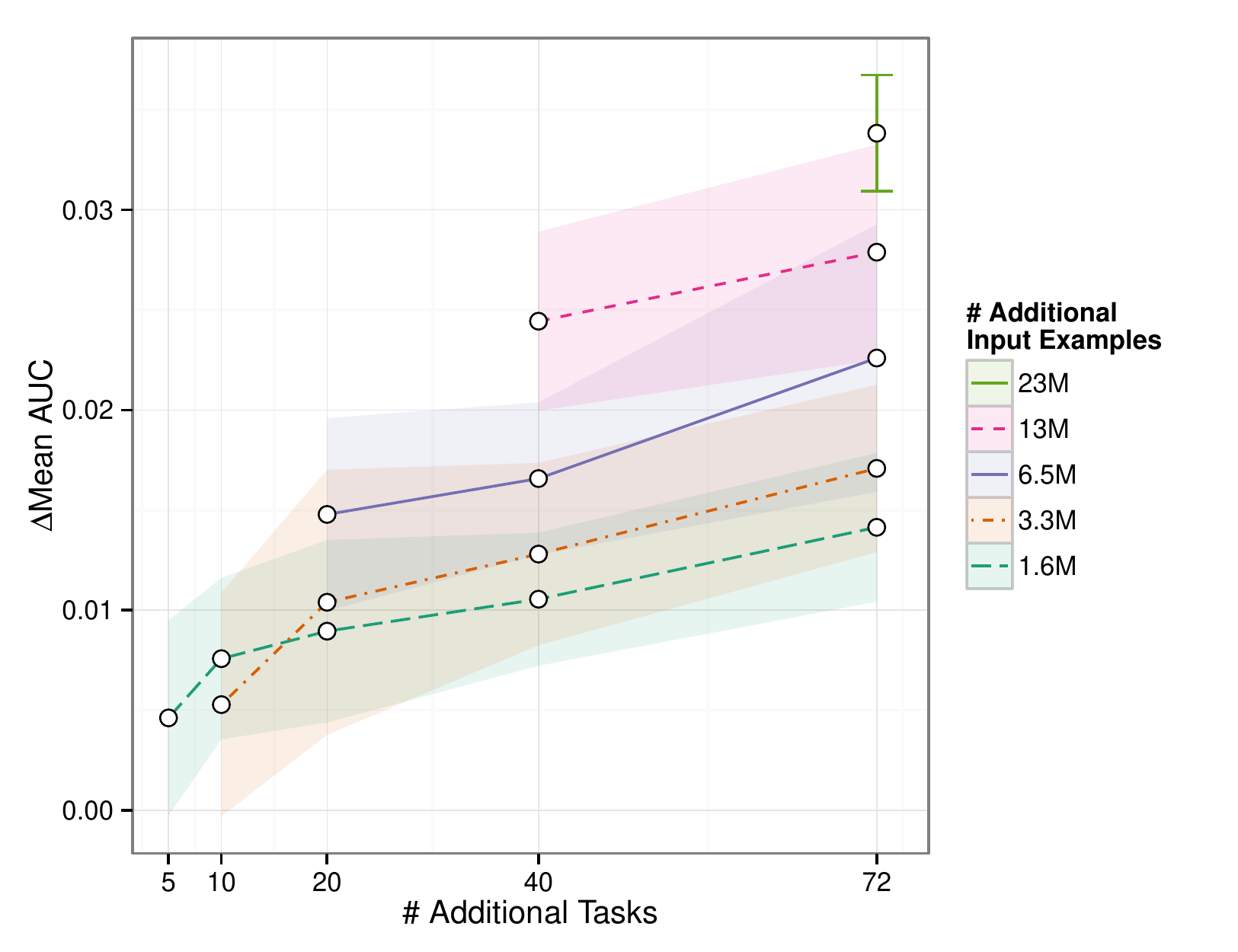}
\caption{Multitask benefit from increasing tasks and data independently.
  As in \figurename~\ref{fig:growth_held_in}, we added randomly selected
  tasks ($x$-axis) to a fixed held-in set. A stratified random sampling
  scheme was applied to the additional tasks in order to achieve fixed
  total numbers of additional input examples (color, line type).  White
  points indicate the mean over $10$ experimental runs of $\Delta$ mean-AUC
  over the initial network trained on the $10$ held-in datasets. Color-filled
  areas and error bars describe the smoothed $95\%$ confidence intervals.}
\label{fig:data_tasks}
\end{figure}

\figurename~\ref{fig:data_tasks} shows the results of our experiments. The
$x$-axis tracks the number of additional tasks, while the $y$-axis displays
the improvement in performance for the held-in set relative to a multitask
network trained only on the held-in data. When the total amount of data is
fixed, having more tasks consistently yields improvement. Similarly, when
the number of tasks is fixed, adding additional data consistently improves
performance. Our results suggest that the total amount of data and the
total number of tasks both contribute significantly to the multitask
effect.

\subsection{Do massively multitask networks extract generalizable features?}
\label{sec:embedding}

The features extracted by the top layer of the network represent
information useful to many tasks. Consequently, we sought to determine the
transferability of these features to tasks not in the training set.  We
held out ten data sets from the growth curves calculated in
Section~\ref{sec:growth_curve} and used the learned weights from points
along the growth curves to initialize single-task networks for the held-out
datasets, which we then fine-tuned.

\begin{figure}[ht]
\vskip 0.2in
\begin{center}
\centerline{\includegraphics[width=\linewidth]{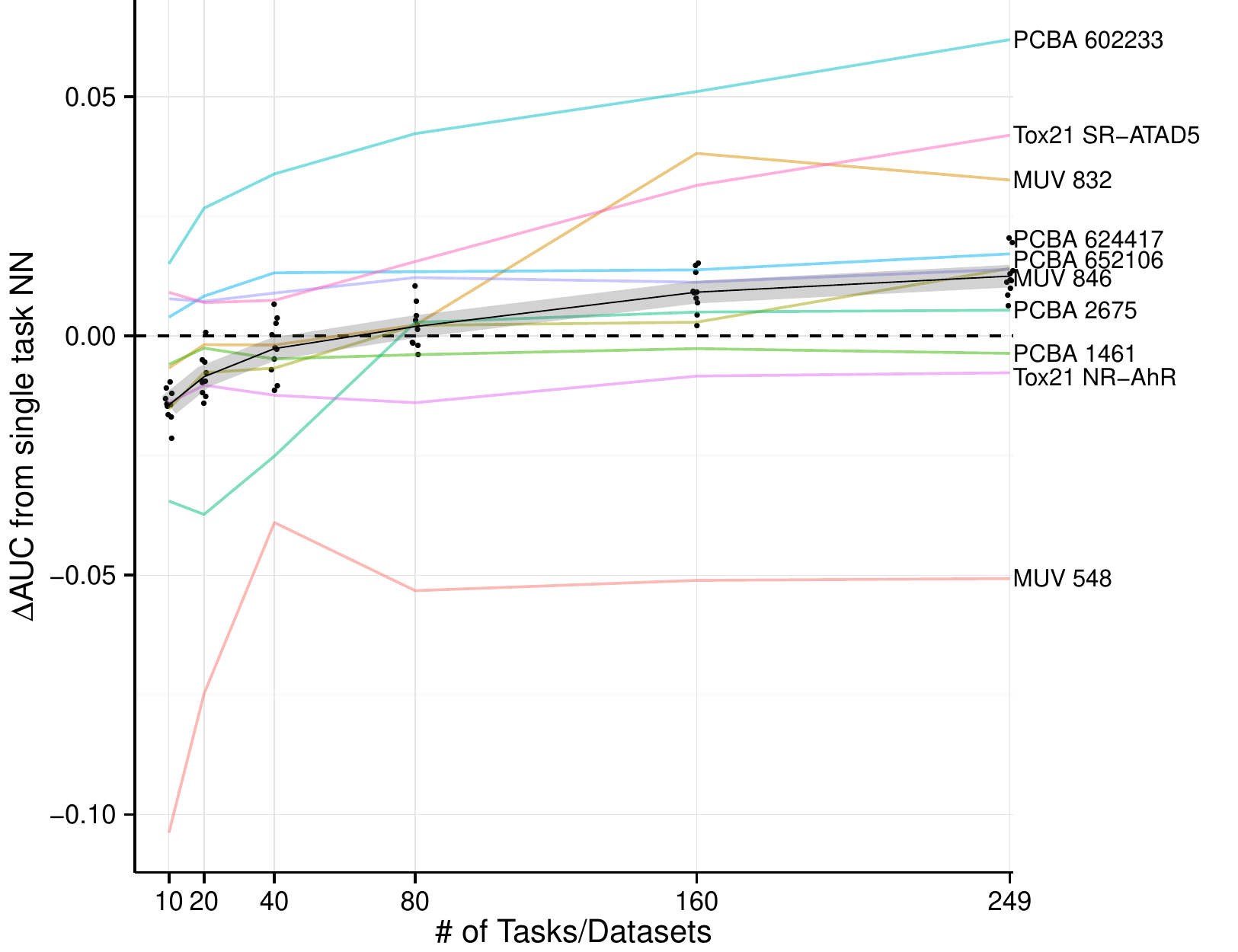}}
\caption{Held-out growth curves. The $y$-axis shows the change in AUC
  compared to a single-task neural network with the same architecture
  (PSTNN). Each colored curve is the result of initializing a single-task
  neural network from the weights of the networks from
  Section~\ref{sec:growth_curve} and computing the mean across the $10$
  experimental runs. These datasets were \emph{not} included in the
  training of the original networks. The shaded curve is the mean across
  the $100$ combinations of datasets and experimental runs, and black dots
  represent means across the $10$ held-out datasets for each experimental
  run, where additional tasks were randomly selected.}
\label{fig:held_out}
\end{center}
\vskip -0.2in
\end{figure}

The results of training these networks (with $5$-fold stratified
cross-validation) are shown in \figurename~\ref{fig:held_out}. First, note
that many of the datasets performed worse than the baseline when
initialized from the 10-held-in-task networks. Further, some datasets never
exhibited any positive effect due to multitask initialization. Transfer
learning can be negative.

Second, note that the transfer learning effect became stronger as multitask
networks were trained on more data. Large multitask networks exhibited
better transferability, but the average effect even with $249$ datasets was
only $\sim.01$ AUC.  We hypothesize that the extent of this
generalizability is determined by the presence or absence of relevant data
in the multitask training set.

\subsection{When do datasets benefit from multitask training?}

The results in Sections \ref{sec:growth_curve} and \ref{sec:embedding}
indicate that some datasets benefit more from multitask training than
others. In an effort to explain these differences, we consider three
specific questions:
\begin{enumerate}
\itemsep0em
\item Do shared active compounds explain multitask improvement?
\item Do some biological target classes realize greater multitask
  improvement than others?
\item Do tasks associated with duplicated targets have artificially high
  multitask performance?
\end{enumerate}

\subsubsection{Shared Active Compounds}
\label{sec:similarity}

The biological context of our datasets implies that active compounds
contain more information than inactive compounds; while an inactive
compound may be inactive for many reasons, active compounds often rely on
similar physical mechanisms. Hence, shared active compounds should be a
good measure of dataset similarity.

\begin{figure}[ht]
\centering
\includegraphics[width=0.9\linewidth]{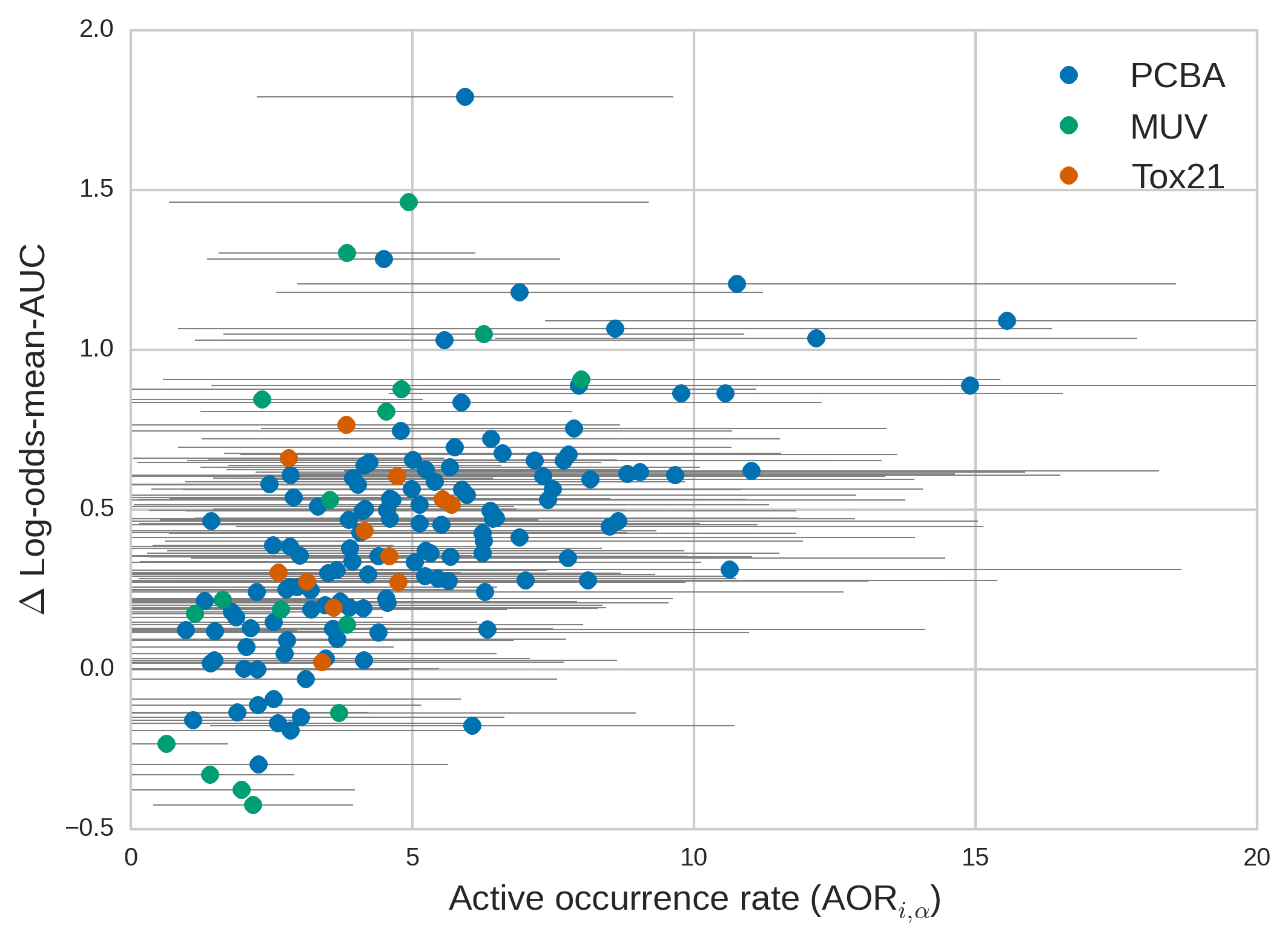}
\caption{Multitask improvement compared to active occurrence rate (AOR).
  Each point in the figure represents a particular dataset $D_i$. The
  $x$-coordinate is the mean AOR across all active compounds in $D_i$, and
  the $y$-coordinate is the difference in log-odds-mean-AUC between
  multitask and single-task models. The gray bars indicate standard
  deviations around the AOR means. There is a moderate correlation
  ($r^2=.33$). For reasons discussed in Section~\ref{sec:experimental}, we
  excluded DUD-E from this analysis. (Including DUD-E results in a similar
  correlation, $r^2=.22$.)}
\label{fig:dataset_similarity}
\end{figure}

\figurename~\ref{fig:dataset_similarity} plots multitask improvement
against a measure of dataset similarity we call ``active occurrence rate''
(AOR). For each active compound $\alpha$ in dataset $D_i$, $\text{AOR}_{i,
\alpha}$ is defined as the number of additional datasets in which this
compound is also active:
\[
\text{AOR}_{i, \alpha} = \sum_{d \neq i} \mathbbm{1}{\left(\alpha \in \text{Actives}\left(D_d\right)\right)}.
\]
Each point in \figurename~\ref{fig:dataset_similarity} corresponds to a
single dataset $D_i$. The $x$-coordinate is
\[
\text{AOR}_i = \Mean_{\alpha \in \text{Actives}\left(D_i\right)}\left(\text{AOR}_{i, \alpha}\right),
\]
and the $y$-coordinate ($\Delta$ log-odds-mean-AUC) is
\[
\small
\text{logit}\left(\frac{1}{K}\sum_{k=1}^{K}\text{AUC}_k^{(M)}\left(D_i\right)\right) - \text{logit}\left(\frac{1}{K}\sum_{k=1}^{K}\text{AUC}_k^{(S)}\left(D_i\right)\right),
\]
where $\text{AUC}_k^{(M)}\left(D_i\right)$ and
$\text{AUC}_k^{(S)}\left(D_i\right)$ are respectively the AUC values for
the $k$-th fold of dataset $i$ in the multitask and single-task models, and
$\text{logit}\left(p\right)=\log\left(p/(1-p)\right)$. The use of log-odds
reduces the effect of outliers and emphasizes changes in AUC when the
baseline is high. Note that for reasons discussed in
Section~\ref{sec:experimental}, DUD-E was excluded from this analysis.

There is a moderate correlation between AOR and $\Delta$ log-odds-mean-AUC
($r^2=.33$); we note that this correlation is not present when we use
$\Delta$ mean-AUC as the $y$-coordinate ($r^2=.09$). We hypothesize that
some portion of the multitask effect is determined by shared active
compounds. That is, a dataset is most likely to benefit from multitask
training when it shares many active compounds with other datasets in the
collection.

\subsubsection{Target Classes}

\begin{figure}[ht]
\centering
\includegraphics[width=0.9\linewidth]{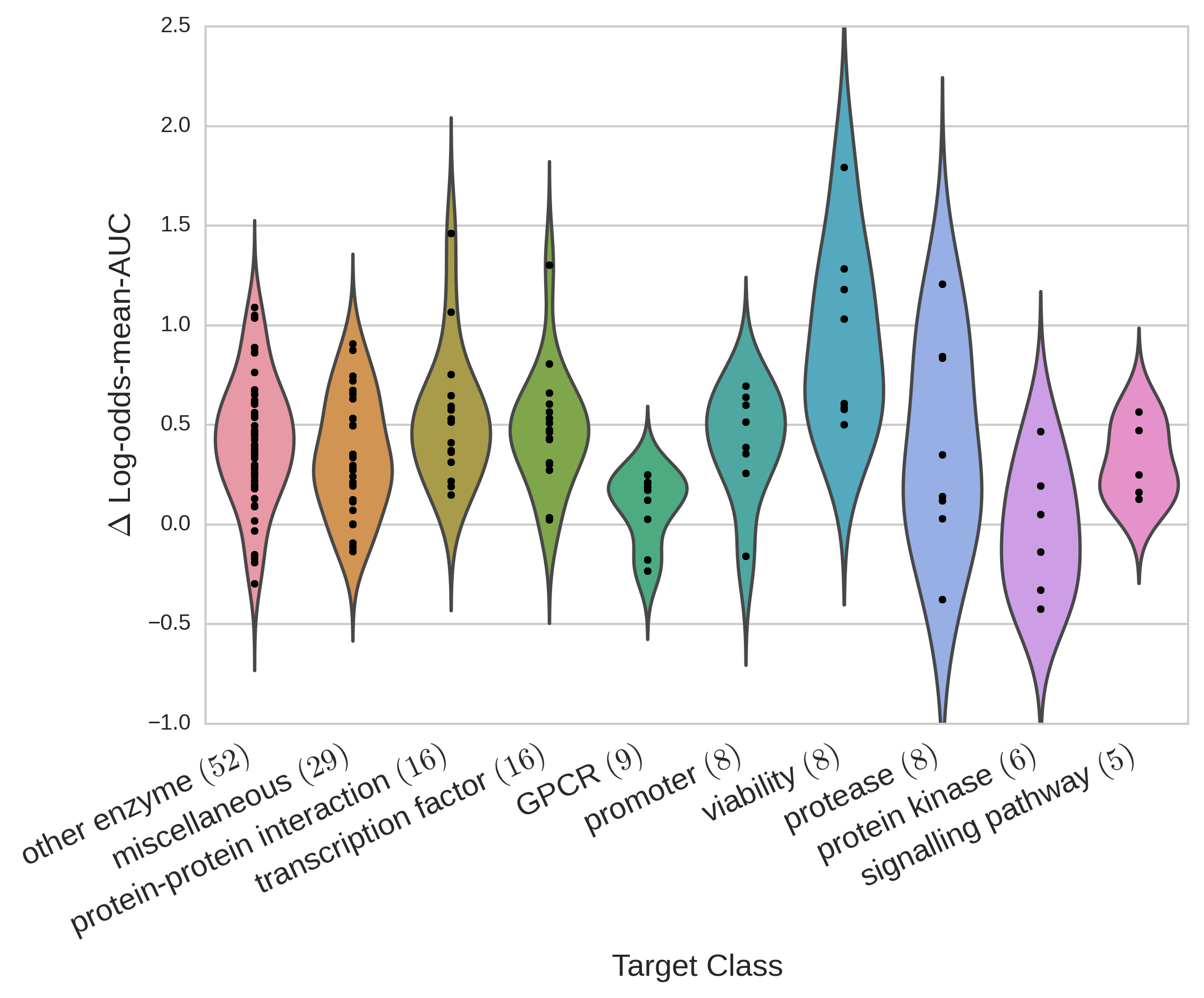}
\caption{Multitask improvement across target classes. The $x$-coordinate
  lists a series of biological target classes represented in our dataset
  collection, and the $y$-coordinate is the difference in log-odds-mean-AUC
  between multitask and single-task models. Note that the DUD-E datasets
  are excluded. Classes are ordered by total number of targets (in
  parenthesis), and target classes with fewer than five members are merged
  into ``miscellaneous.''}
\label{fig:target_similarity}
\end{figure}

\figurename~\ref{fig:target_similarity} shows the relationship between
multitask improvement and target classes. As before, we report multitask
improvement in terms of log-odds and exclude the DUD-E datasets.
Qualitatively, no target class benefited more than any other from multitask
training. Nearly every target class realized gains, suggesting that the
multitask framework is applicable to experimental data from multiple target
classes.

\subsubsection{Duplicate Targets}
\label{sec:duplicates}

As mentioned in Section~\ref{sec:datasets}, there are many cases of tasks
with identical targets. We compared the multitask improvement of duplicate
vs. unique tasks. The distributions have substantial overlap (see the
Appendix), but the average log-odds improvement was slightly higher for
duplicated tasks ($.531$ vs. $.372$; a one-sided $t$-test between the
duplicate and unique distributions gave $p=.016$). Since duplicated targets
are likely to share many active compounds, this improvement is consistent
with the correlation seen in Section~\ref{sec:similarity}. However, sign
tests for single-task vs.  multitask models for duplicate and unique
targets gave significant and highly overlapping confidence intervals
($[0.04, 0.24]$ and $[0.06, 0.17]$, respectively; recall that the meaning
of these intervals is given in the caption for
\tablename~\ref{tab:exp_results}). Together, these results suggest that
there is not significant information leakage within multitask networks.
Consequently, the results of our analysis are unlikely to be significantly
affected by the presence of duplicate targets in our dataset collection.

\section{Discussion and Conclusion}
In this work, we investigated the use of massively multitask networks for
virtual screening. We gathered a large collection of publicly available
experimental data that we used to train massively multitask neural
networks. These networks achieved significant improvement over simple
machine learning algorithms.

We explored several aspects of the multitask framework. First, we
demonstrated that multitask performance improved with the addition of more
tasks; our performance was still climbing at 259 tasks. Next, we considered
the relative importance of introducing more data vs. more tasks. We found
that additional data and additional tasks both contributed significantly to
the multitask effect. We next discovered that multitask learning afforded
limited transferability to tasks not contained in the training set. This
effect was not universal, and required large amounts of data even when it
did apply.

We observed that the multitask effect was stronger for some datasets than
others. Consequently, we investigated possible explanations for this
discrepancy and found that the presence of shared active compounds was
moderately correlated with multitask improvement, but the biological class
of the target was not. It is also possible that multitask improvement
results from accurately modeling experimental artifacts rather than
specific interactions between targets and small molecules. We do not
believe this to be the case, as we demonstrated strong improvement on the
thoroughly-cleaned MUV datasets.

The efficacy of multitask learning is directly related to the availability
of relevant data. Hence, obtaining greater amounts of data is of critical
importance for improving the state of the art. Major pharmaceutical
companies possess vast private stores of experimental measurements; our
work provides a strong argument that increased data sharing could result in
benefits for all.

More data will maximize the benefits achievable using current
architectures, but in order for algorithmic progress to occur, it must be
possible to judge the performance of proposed models against previous work.
It is disappointing to note that all published applications of deep
learning to virtual screening (that we are aware of) use distinct datasets
that are not directly comparable. It remains to future research to
establish standard datasets and performance metrics for this field.

Another direction for future work is the further study of small molecule
featurization. In this work, we use only one possible featurization
(ECFP4), but there exist many others. Additional performance may also be
realized by considering targets as well as small molecules in the
featurization. Yet another line of research could improve performance by
using unsupervised learning to explore much larger segments of chemical
space.

Although deep learning offers interesting possibilities for virtual
screening, the full drug discovery process remains immensely complicated.
Can deep learning---coupled with large amounts of experimental
data---trigger a revolution in this field? Considering the transformational
effect that these methods have had on other fields, we are optimistic about
the future.

\section*{Acknowledgments} B.R. was supported by the Fannie and John Hertz
Foundation.  S.K. was supported by a Smith Stanford Graduate Fellowship.
We also acknowledge support from NIH and NSF, in particular NIH U54
GM072970 and NSF 0960306. The latter award was funded under the American
Recovery and Reinvestment Act of 2009 (Public Law 111-5).
\putbib
\end{bibunit}

\begin{bibunit}
\onecolumn
\pagenumbering{gobble}
\renewcommand{\thefigure}{A\arabic{figure}}
\renewcommand{\thetable}{A\arabic{table}}
\renewcommand\thesection{\Alph{section}}
\renewcommand\thepart{\Alph{part}}
\counterwithin{figure}{section}
\counterwithin{table}{section}
\setcounter{section}{0}
\maketitle
\section{Dataset Construction and Design}

The PCBA datasets are dose-response assays performed by the NCATS Chemical
Genomics Center (NCGC) and downloaded from PubChem BioAssay using the
following search limits: TotalSidCount from 10000, ActiveSidCount from 30,
Chemical, Confirmatory, Dose-Response, Target: Single, NCGC. These limits
correspond to the search query: (10000[TotalSidCount] :
1000000000[TotalSidCount]) AND (30[ActiveSidCount] :
1000000000[ActiveSidCount]) AND ``small\_molecule''[filt] AND
``doseresponse''[filt] AND 1[TargetCount] AND ``NCGC''[SourceName].  We
note that the DUD-E datasets are especially susceptible to ``artificial
enrichment'' (unrealistic divisions between active and inactive compounds)
as an artifact of the dataset construction procedure. Each data point in
our collection was associated with a binary label classifying it as either
active or inactive.

\begin{figure}[ht]
\centering
\includegraphics[width=\linewidth]{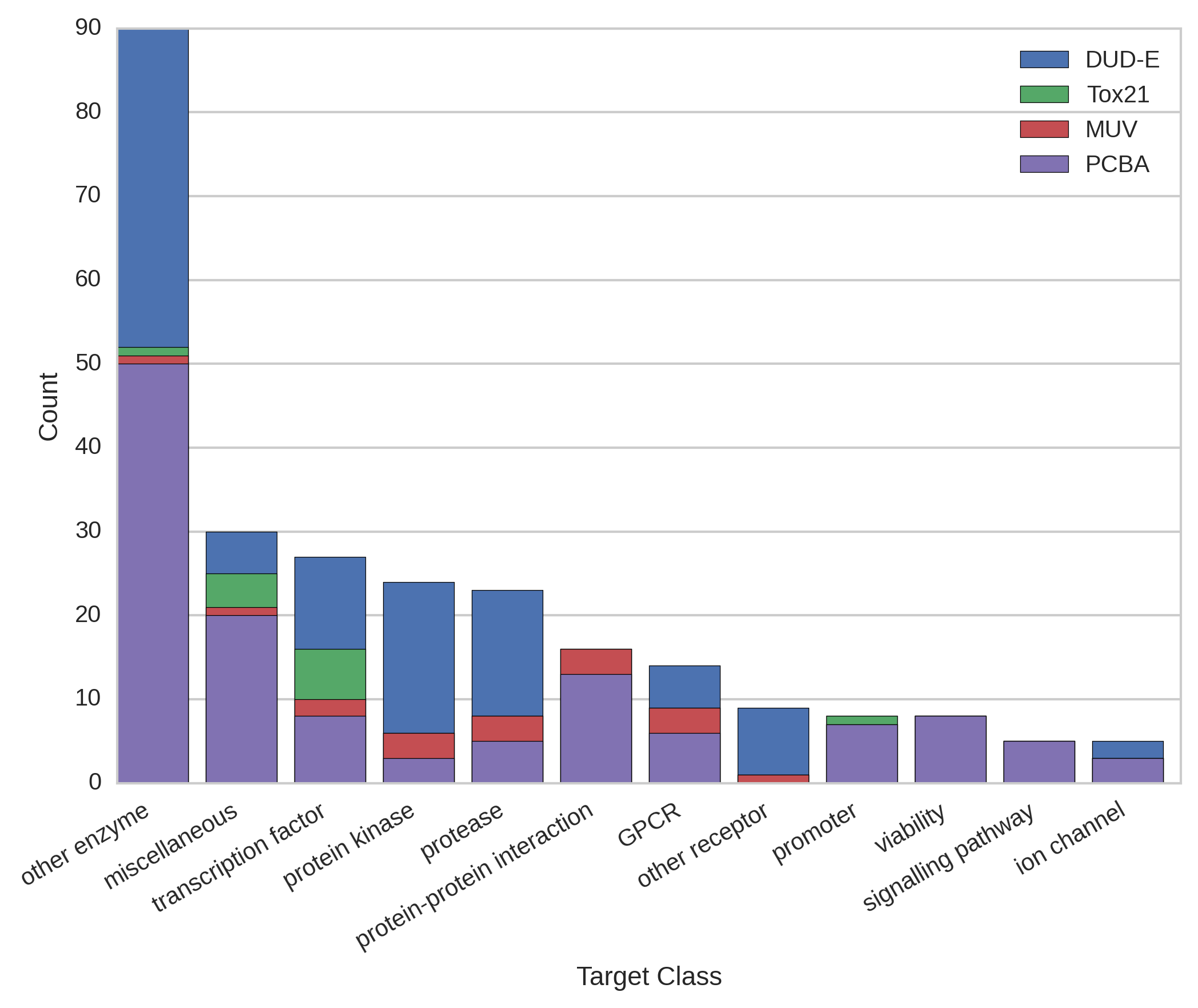}
\caption{Target class breakdown. Classes with fewer than five members were
  merged into the ``miscellaneous'' class.}
\label{fig:target_bar}
\end{figure}

A description of each of our 259 datasets is given in \tablename~A1.  These
datasets cover a wide range of target classes and assay types, including
both cell-based and in vitro experiments. Datasets with duplicated targets
are marked with an asterisk (note that only the non-DUD-E duplicate target
datasets were used in the analysis described in the text). For the PCBA
datasets, compounds not labeled ``Active'' were considered inactive
(including compounds marked ``Inconclusive''). Due to missing data in
PubChem BioAssay and/or featurization errors, some data points and
compounds were not used for evaluation of our models; failure rates for
each dataset group are shown in \tablename~\ref{tab:failures}. The Tox21
group suffered especially high failure rates, likely due to the relatively
large number of metallic or otherwise abnormal compounds that are not
supported by the RDKit package.  The counts given in \tablename~A1 do not
include these missing data. A graphical breakdown of the datasets by target
class is shown in \figurename~\ref{fig:target_bar}. The datasets used for
the held-in and held-out analyses are repeated in
\tablename~\ref{tab:held-in} and \tablename~\ref{tab:held-out},
respectively.

\begin{figure}[ht]
\centering
\includegraphics[width=\linewidth]{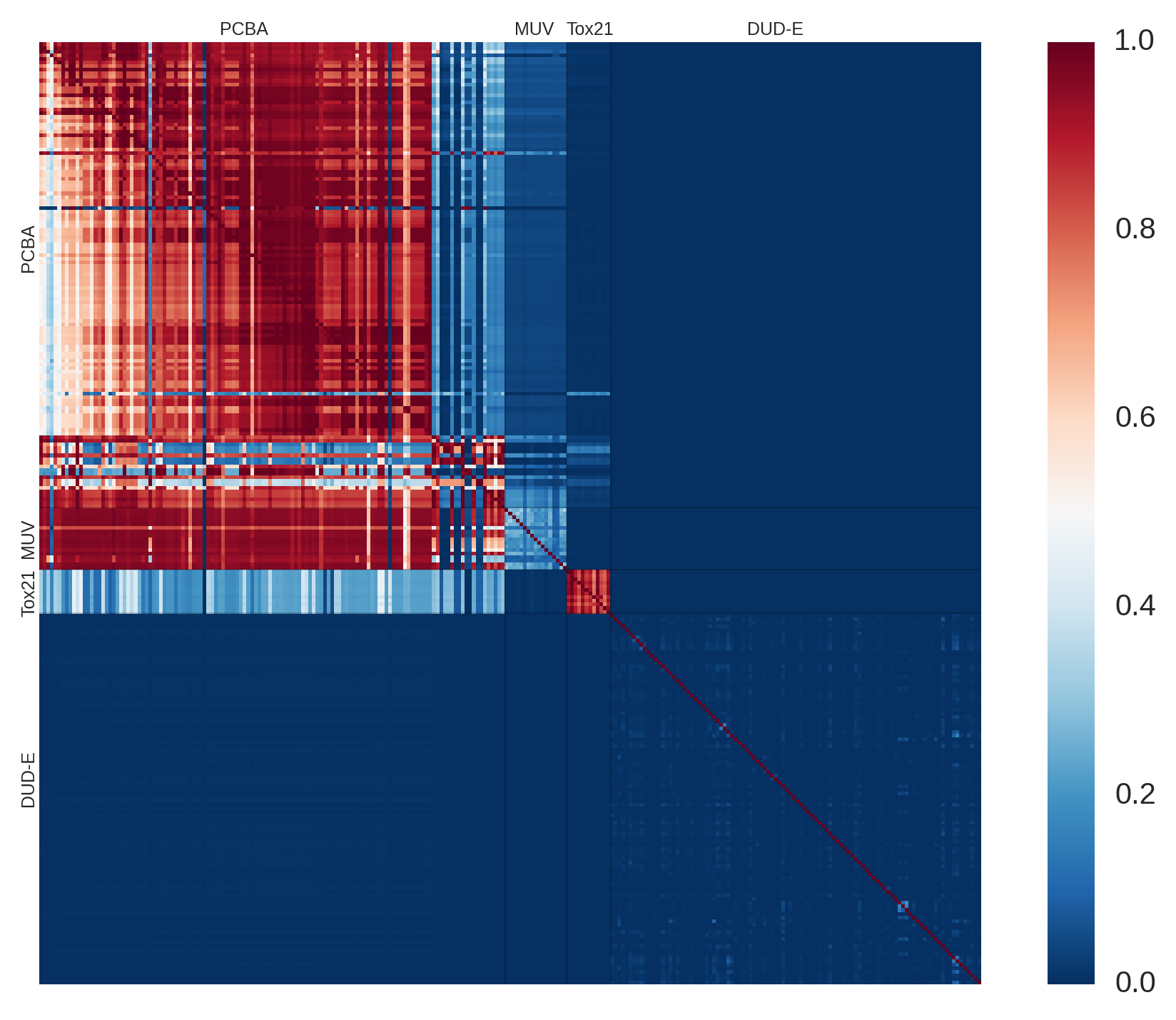}
\caption{Pairwise dataset intersections. The value of the element at
  position $(x, y)$ corresponds to the fraction of dataset $x$ that is
  contained in dataset $y$. Thin black lines are used to indicate divisions
  between dataset groups.}
\label{fig:dataset_heatmap}
\end{figure}

As an extension of our treatment of task similarity in the text, we
generated the heatmap in \figurename~\ref{fig:dataset_heatmap} to show the
pairwise intersection between all datasets in our collection. A few
characteristics of our datasets are immediately apparent: \begin{itemize}
\item The datasets in the DUD-E group have very little intersection with
any other datasets.  \item The PCBA and Tox21 datasets have substantial
self-overlap. In contrast, the MUV datasets have relatively little
self-overlap.  \item The MUV datasets have substantial overlap with the
datasets in the PCBA group.  \item The Tox21 datasets have very small
intersections with datasets in other groups.  \end{itemize}

\begin{figure}[ht]
\centering
\includegraphics[width=0.5\linewidth]{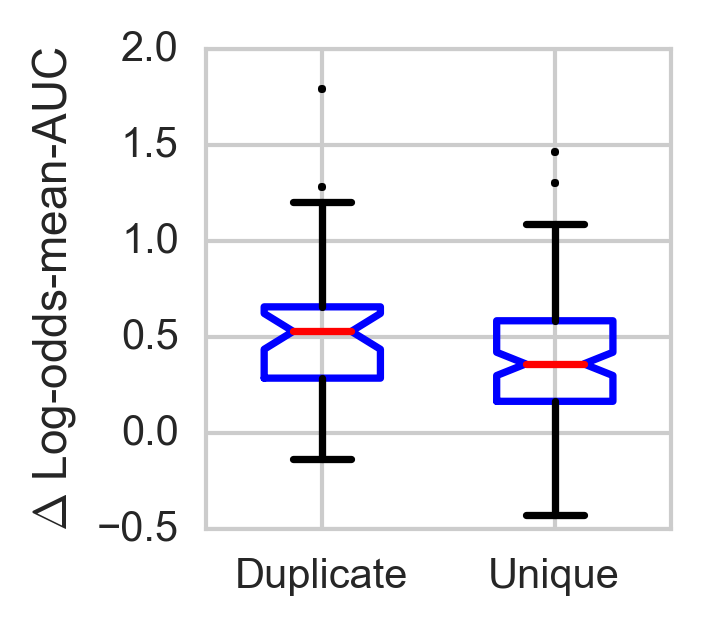}
\caption{Multitask performance of duplicate and unique targets. Outliers
  are omitted for clarity. Notches indicate a confidence interval around the
  median, computed as $\pm 1.57 \times \text{IQR}/ \sqrt{N}$
  \citep{mcgill1978variations}.}
\label{fig:duplicates}
\end{figure}

\figurename~\ref{fig:duplicates} shows the $\Delta$ log-odds-mean-AUC for
datasets with duplicate and unique targets.

\rowcolors{2}{gray!25}{white}
\begin{filecontents*}{datasets.csv}
Dataset,Actives,Inactives,Target Class,Target
pcba-aid411*,1562,69734,other enzyme,luciferase
pcba-aid875,32,73870,protein-protein interaction,brca1-bach1
pcba-aid881,589,106656,other enzyme,15hLO-2
pcba-aid883,1214,8170,other enzyme,CYP2C9
pcba-aid884,3391,9676,other enzyme,CYP3A4
pcba-aid885,163,12904,other enzyme,CYP3A4
pcba-aid887,1024,72140,other enzyme,15hLO-1
pcba-aid891,1548,7836,other enzyme,CYP2D6
pcba-aid899,1809,7575,other enzyme,CYP2C19
pcba-aid902*,1872,123512,viability,H1299-p53A138V
pcba-aid903*,338,54175,viability,H1299-neo
pcba-aid904*,528,53981,viability,H1299-neo
pcba-aid912,445,68506,miscellaneous,anthrax LF-PA internalization
pcba-aid914,218,10619,transcription factor,HIF-1
pcba-aid915,436,10401,transcription factor,HIF-1
pcba-aid924*,1146,122867,viability,H1299-p53A138V
pcba-aid925,39,64358,miscellaneous,EGFP-654
pcba-aid926,350,71666,GPCR,TSHR
pcba-aid927*,61,59108,protease,USP2a
pcba-aid938,1775,70241,ion channel,CNG
pcba-aid995*,699,70189,signalling pathway,ERK1/2 cascade
pcba-aid1030,15963,200920,other enzyme,ALDH1A1
pcba-aid1379*,562,198500,other enzyme,luciferase
pcba-aid1452,177,151634,other enzyme,12hLO
pcba-aid1454*,536,130788,signalling pathway,ERK1/2 cascade
pcba-aid1457,722,204859,other enzyme,IMPase
pcba-aid1458,5805,202680,miscellaneous,SMN2
pcba-aid1460*,5662,261757,protein-protein interaction,K18
pcba-aid1461,2305,218561,GPCR,NPSR
pcba-aid1468*,1039,270371,protein-protein interaction,K18
pcba-aid1469,169,276098,protein-protein interaction,TRb-SRC2
pcba-aid1471,288,223321,protein-protein interaction,huntingtin
pcba-aid1479,788,275479,miscellaneous,TRb-SRC2
pcba-aid1631,892,262774,other enzyme,hPK-M2
pcba-aid1634,154,263512,other enzyme,hPK-M2
pcba-aid1688,2374,218200,protein-protein interaction,HTTQ103
pcba-aid1721,1087,291649,other enzyme,LmPK
pcba-aid2100*,1159,301145,other enzyme,alpha-glucosidase
pcba-aid2101*,285,321268,other enzyme,glucocerebrosidase
pcba-aid2147,3477,223441,other enzyme,JMJD2E
pcba-aid2242*,715,198459,other enzyme,alpha-glucosidase
pcba-aid2326,1069,268500,miscellaneous,influenza A NS1
pcba-aid2451,2008,285737,other enzyme,FBPA
pcba-aid2517,1136,344762,other enzyme,APE1
pcba-aid2528,660,347283,other enzyme,BLM
pcba-aid2546,10550,293509,transcription factor,VP16
pcba-aid2549,1210,233706,other enzyme,RECQ1
pcba-aid2551,16666,288772,transcription factor,ROR gamma
pcba-aid2662,110,293953,miscellaneous,MLL-HOX-A
pcba-aid2675,99,279333,miscellaneous,MBNL1-CUG
pcba-aid2676,1081,361124,GPCR,RXFP1
pcba-aid463254*,41,330640,protease,USP2a
pcba-aid485281,254,341253,miscellaneous,apoferritin
pcba-aid485290,942,343503,other enzyme,TDP1
pcba-aid485294*,148,362056,other enzyme,AmpC
pcba-aid485297,9126,311481,promoter,Rab9
pcba-aid485313,7567,313119,promoter,NPC1
pcba-aid485314,4491,329974,other enzyme,DNA polymerase beta
pcba-aid485341*,1729,328952,other enzyme,AmpC
pcba-aid485349,618,321745,protein kinase,ATM
pcba-aid485353,603,328042,protease,PLP
pcba-aid485360,1485,223830,protein-protein interaction,L3MBTL1
pcba-aid485364,10700,345950,other enzyme,TGR
pcba-aid485367,557,330124,other enzyme,PFK
pcba-aid492947,80,330601,GPCR,beta2-AR
pcba-aid493208,342,43647,protein kinase,mTOR
pcba-aid504327,759,380820,other enzyme,GCN5L2
pcba-aid504332,30586,317753,other enzyme,G9a
pcba-aid504333,15670,341165,protein-protein interaction,BAZ2B
pcba-aid504339,16857,367661,protein-protein interaction,JMJD2A
pcba-aid504444,7390,353475,transcription factor,Nrf2
pcba-aid504466,4169,325944,viability,HEK293T-ELG1-luc
pcba-aid504467,7647,322464,promoter,ELG1
pcba-aid504706,201,321230,miscellaneous,p53
pcba-aid504842,101,329517,other enzyme,Mm-CPN
pcba-aid504845,104,385400,miscellaneous,RGS4
pcba-aid504847,3515,390525,transcription factor,VDR
pcba-aid504891,34,383652,other enzyme,Pin1
pcba-aid540276*,4494,279673,miscellaneous,Marburg virus
pcba-aid540317,2126,381226,protein-protein interaction,HP1-beta
pcba-aid588342*,25034,335826,other enzyme,luciferase
pcba-aid588453*,3921,382731,other enzyme,TrxR1
pcba-aid588456*,51,386206,other enzyme,TrxR1
pcba-aid588579,1987,393298,other enzyme,DNA polymerase kappa
pcba-aid588590,3936,382117,other enzyme,DNA polymerase iota
pcba-aid588591,4715,383994,other enzyme,DNA polymerase eta
pcba-aid588795,1308,384951,other enzyme,FEN1
pcba-aid588855,4894,398438,transcription factor,Smad3
pcba-aid602179,364,387230,other enzyme,IDH1
pcba-aid602233,165,380904,other enzyme,PGK
pcba-aid602310,310,402026,protein-protein interaction,Vif-A3G
pcba-aid602313,762,383076,protein-protein interaction,Vif-A3F
pcba-aid602332,70,415773,promoter,GRP78
pcba-aid624170,837,404440,other enzyme,GLS
pcba-aid624171,1239,402621,transcription factor,Nrf2
pcba-aid624173,488,406224,other enzyme,PYK
pcba-aid624202,3968,372045,promoter,BRCA1
pcba-aid624246,101,367273,miscellaneous,ERG
pcba-aid624287,423,334388,signalling pathway,Gsgsp
pcba-aid624288,1356,336077,signalling pathway,Gsgsp
pcba-aid624291,222,345619,promoter,a7
pcba-aid624296*,9841,333378,miscellaneous,DNA re-replication
pcba-aid624297*,6214,336050,miscellaneous,DNA re-replication
pcba-aid624417,6388,398731,GPCR,GLP-1
pcba-aid651635,3784,387779,promoter,ATXN
pcba-aid651644,748,361115,miscellaneous,Vpr
pcba-aid651768,1677,362320,other enzyme,WRN
pcba-aid651965,6422,331953,protease,ClpP
pcba-aid652025,238,364365,signalling pathway,IL-2
pcba-aid652104,7126,396566,miscellaneous,TDP-43
pcba-aid652105,4072,324774,other enzyme,PI5P4K
pcba-aid652106,496,368281,miscellaneous,alpha-synuclein
pcba-aid686970,5949,358501,viability,HT-1080-NT
pcba-aid686978*,62746,354086,viability,DT40-hTDP1
pcba-aid686979*,48816,368048,viability,DT40-hTDP1
pcba-aid720504,10170,353881,protein kinase,Plk1 PBD
pcba-aid720532*,945,14532,miscellaneous,Marburg virus
pcba-aid720542,733,363349,protein-protein interaction,AMA1-RON2
pcba-aid720551*,1265,342387,ion channel,KCHN2 3.1
pcba-aid720553*,3260,338810,ion channel,KCHN2 3.1
pcba-aid720579*,1913,304815,miscellaneous,orthopoxvirus
pcba-aid720580*,1508,324844,miscellaneous,orthopoxvirus
pcba-aid720707,268,364332,other enzyme,EPAC1
pcba-aid720708,661,363939,other enzyme,EPAC2
pcba-aid720709,516,364084,other enzyme,EPAC1
pcba-aid720711,290,364310,other enzyme,EPAC2
pcba-aid743255,902,388656,protease,USP1/UAF1
pcba-aid743266,306,405368,GPCR,PTHR1
muv-aid466,30,14999,GPCR,S1P1 receptor
muv-aid548,30,15000,protein kinase,PKA
muv-aid600,30,14999,transcription factor,SF1
muv-aid644,30,14998,protein kinase,Rho-Kinase2
muv-aid652,30,15000,other enzyme,HIV RT-RNase
muv-aid689,30,14999,other receptor,Eph rec. A4
muv-aid692,30,15000,transcription factor,SF1
muv-aid712*,30,14997,miscellaneous,HSP90
muv-aid713*,30,15000,protein-protein interaction,ER-a-coact. bind.
muv-aid733,30,15000,protein-protein interaction,ER-b-coact. bind.
muv-aid737*,30,14999,protein-protein interaction,ER-a-coact. bind.
muv-aid810*,30,14999,protein kinase,FAK
muv-aid832,30,15000,protease,Cathepsin G
muv-aid846,30,15000,protease,FXIa
muv-aid852,30,15000,protease,FXIIa
muv-aid858,30,14999,GPCR,D1 receptor
muv-aid859,30,15000,GPCR,M1 receptor
tox-NR-AhR,768,5780,transcription factor,Aryl hydrocarbon receptor
tox-NR-AR-LBD*,237,6520,transcription factor,Androgen receptor
tox-NR-AR*,309,6955,transcription factor,Androgen receptor
tox-NR-Aromatase,300,5521,other enzyme,Aromatase
tox-NR-ER-LBD*,350,6604,transcription factor,Estrogen receptor alpha
tox-NR-ER*,793,5399,transcription factor,Estrogen receptor alpha
tox-NR-PPAR-gamma*,186,6263,transcription factor,PPARg
tox-SR-ARE,942,4889,miscellaneous,ARE
tox-SR-ATAD5,264,6807,promoter,ATAD5
tox-SR-HSE,372,6094,miscellaneous,HSE
tox-SR-MMP,919,4891,miscellaneous,mitochondrial membrane potential
tox-SR-p53,423,6351,miscellaneous,p53 signalling
dude-aa2ar,482,31546,GPCR,Adenosine A2a receptor
dude-abl1,182,10749,protein kinase,Tyrosine-protein kinase ABL
dude-ace,282,16899,protease,Angiotensin-converting enzyme
dude-aces,453,26240,other enzyme,Acetylcholinesterase
dude-ada,93,5450,other enzyme,Adenosine deaminase
dude-ada17,532,35900,protease,ADAM17
dude-adrb1,247,15848,GPCR,Beta-1 adrenergic receptor
dude-adrb2,231,14997,GPCR,Beta-2 adrenergic receptor
dude-akt1,293,16441,protein kinase,Serine/threonine-protein kinase AKT
dude-akt2,117,6899,protein kinase,Serine/threonine-protein kinase AKT2
dude-aldr,159,8999,other enzyme,Aldose reductase
dude-ampc,48,2850,other enzyme,Beta-lactamase
dude-andr*,269,14350,transcription factor,Androgen Receptor
dude-aofb,122,6900,other enzyme,Monoamine oxidase B
dude-bace1,283,18097,protease,Beta-secretase 1
dude-braf,152,9950,protein kinase,Serine/threonine-protein kinase B-raf
dude-cah2,492,31168,other enzyme,Carbonic anhydrase II
dude-casp3,199,10700,protease,Caspase-3
dude-cdk2,474,27850,protein kinase,Cyclin-dependent kinase 2
dude-comt,41,3850,other enzyme,Catechol O-methyltransferase
dude-cp2c9,120,7449,other enzyme,Cytochrome P450 2C9
dude-cp3a4,170,11800,other enzyme,Cytochrome P450 3A4
dude-csf1r,166,12149,other receptor,Macrophage colony stimulating factor receptor
dude-cxcr4,40,3406,GPCR,C-X-C chemokine receptor type 4
dude-def,102,5700,other enzyme,Peptide deformylase
dude-dhi1,330,19350,other enzyme,11-beta-hydroxysteroid dehydrogenase 1
dude-dpp4,533,40943,protease,Dipeptidyl peptidase IV
dude-drd3,480,34037,GPCR,Dopamine D3 receptor
dude-dyr,231,17192,other enzyme,Dihydrofolate reductase
dude-egfr,542,35047,other receptor,Epidermal growth factor receptor erbB1
dude-esr1*,383,20675,transcription factor,Estrogen receptor alpha
dude-esr2,367,20190,transcription factor,Estrogen receptor beta
dude-fa10,537,28315,protease,Coagulation factor X
dude-fa7,114,6250,protease,Coagulation factor VII
dude-fabp4,47,2750,miscellaneous,Fatty acid binding protein adipocyte
dude-fak1*,100,5350,protein kinase,FAK
dude-fgfr1,139,8697,other receptor,Fibroblast growth factor receptor 1
dude-fkb1a,111,5800,other enzyme,FK506-binding protein 1A
dude-fnta,592,51481,other enzyme,Protein farnesyltransferase/geranylgeranyltransferase type I alpha subunit
dude-fpps,85,8829,other enzyme,Farnesyl diphosphate synthase
dude-gcr,258,14999,transcription factor,Glucocorticoid receptor
dude-glcm*,54,3800,other enzyme,glucocerebrosidase
dude-gria2,158,11842,ion channel,Glutamate receptor ionotropic
dude-grik1,101,6549,ion channel,Glutamate receptor ionotropic kainate 1
dude-hdac2,185,10299,other enzyme,Histone deacetylase 2
dude-hdac8,170,10449,other enzyme,Histone deacetylase 8
dude-hivint,100,6650,other enzyme,Human immunodeficiency virus type 1 integrase
dude-hivpr,536,35746,protease,Human immunodeficiency virus type 1 protease
dude-hivrt,338,18891,other enzyme,Human immunodeficiency virus type 1 reverse transcriptase
dude-hmdh,170,8748,other enzyme,HMG-CoA reductase
dude-hs90a*,88,4849,miscellaneous,HSP90
dude-hxk4,92,4700,other enzyme,Hexokinase type IV
dude-igf1r,148,9298,other receptor,Insulin-like growth factor I receptor
dude-inha,43,2300,other enzyme,Enoyl-[acyl-carrier-protein] reductase
dude-ital,138,8498,miscellaneous,Leukocyte adhesion glycoprotein LFA-1 alpha
dude-jak2,107,6499,protein kinase,Tyrosine-protein kinase JAK2
dude-kif11,116,6849,miscellaneous,Kinesin-like protein 1
dude-kit,166,10449,other receptor,Stem cell growth factor receptor
dude-kith,57,2849,other enzyme,Thymidine kinase
dude-kpcb,135,8700,protein kinase,Protein kinase C beta
dude-lck,420,27397,protein kinase,Tyrosine-protein kinase LCK
dude-lkha4,171,9450,protease,Leukotriene A4 hydrolase
dude-mapk2,101,6150,protein kinase,MAP kinase-activated protein kinase 2
dude-mcr,94,5150,transcription factor,Mineralocorticoid receptor
dude-met,166,11247,other receptor,Hepatocyte growth factor receptor
dude-mk01,79,4549,protein kinase,MAP kinase ERK2
dude-mk10,104,6600,protein kinase,c-Jun N-terminal kinase 3
dude-mk14,578,35848,protein kinase,MAP kinase p38 alpha
dude-mmp13,572,37195,protease,Matrix metalloproteinase 13
dude-mp2k1,121,8149,protein kinase,Dual specificity mitogen-activated protein kinase kinase 1
dude-nos1,100,8048,other enzyme,Nitric-oxide synthase
dude-nram,98,6199,other enzyme,Neuraminidase
dude-pa2ga,99,5150,other enzyme,Phospholipase A2 group IIA
dude-parp1,508,30049,other enzyme,Poly [ADP-ribose] polymerase-1
dude-pde5a,398,27547,other enzyme,Phosphodiesterase 5A
dude-pgh1,195,10800,other enzyme,Cyclooxygenase-1
dude-pgh2,435,23149,other enzyme,Cyclooxygenase-2
dude-plk1,107,6800,protein kinase,Serine/threonine-protein kinase PLK1
dude-pnph,103,6950,other enzyme,Purine nucleoside phosphorylase
dude-ppara,373,19397,transcription factor,PPARa
dude-ppard,240,12247,transcription factor,PPARd
dude-pparg*,484,25296,transcription factor,PPARg
dude-prgr,293,15648,transcription factor,Progesterone receptor
dude-ptn1,130,7250,other enzyme,Protein-tyrosine phosphatase 1B
dude-pur2,50,2698,other enzyme,GAR transformylase
dude-pygm,77,3948,other enzyme,Muscle glycogen phosphorylase
dude-pyrd,111,6450,other enzyme,Dihydroorotate dehydrogenase
dude-reni,104,6958,protease,Renin
dude-rock1,100,6299,protein kinase,Rho-associated protein kinase 1
dude-rxra,131,6948,transcription factor,Retinoid X receptor alpha
dude-sahh,63,3450,other enzyme,Adenosylhomocysteinase
dude-src,524,34491,protein kinase,Tyrosine-protein kinase SRC
dude-tgfr1,133,8500,other receptor,TGF-beta receptor type I
dude-thb,103,7448,transcription factor,Thyroid hormone receptor beta-1
dude-thrb,461,26999,protease,Thrombin
dude-try1,449,25967,protease,Trypsin I
dude-tryb1,148,7648,protease,Tryptase beta-1
dude-tysy,109,6748,other enzyme,Thymidylate synthase
dude-urok,162,9850,protease,Urokinase-type plasminogen activator
dude-vgfr2,409,24946,other receptor,Vascular endothelial growth factor receptor 2
dude-wee1,102,6150,protein kinase,Serine/threonine-protein kinase WEE1
dude-xiap,100,5149,miscellaneous,Inhibitor of apoptosis protein 3
\end{filecontents*}
\csvreader[
longtable=lSSp{1in}p{1.9in},
table head={\toprule\bfseries Dataset & \bfseries Actives & \bfseries Inactives & \bfseries Target Class & \bfseries Target \\
\midrule\endhead\bottomrule\endfoot},
]
{datasets.csv}{1=\Dataset, 2=\Actives, 3=\Inactives, 4=\Class, 5=\Target}{\Dataset & \Actives & \Inactives & \Class & \Target}

\begin{table}[ht]
\centering
\caption{Featurization failures.}
\label{tab:failures}
\vskip 0.2in
\begin{tabular}{lSSS}
\toprule
Group & \text{Original} & \text{Featurized} & \text{Failure Rate (\%)} \\
\midrule
PCBA & 439879 & 437928 & 0.44 \\
DUD-E & 1200966 & 1200406 & 0.05 \\ 
MUV & 95916 & 95899 & 0.02 \\
Tox21 & 11764 & 7830 & 33.44 \\
\bottomrule
\end{tabular}
\end{table}

\clearpage

\begin{table}[ht]
\centering
\caption{Held-in datasets.}
\label{tab:held-in}
\vskip 0.2in
\begin{tabular}{lSSll}
\toprule
\bfseries Dataset & \bfseries Actives & \bfseries Inactives & \bfseries Target Class & \bfseries Target \\
\midrule
pcba-aid899 & 1809 & 7575 & other enzyme & CYP2C19 \\
pcba-aid485297 & 9126 & 311481 & promoter & Rab9 \\
pcba-aid651644 & 748 & 361115 & miscellaneous & Vpr \\
pcba-aid651768 & 1677 & 362320 & other enzyme & WRN \\
pcba-aid743266 & 306 & 405368 & GPCR & PTHR1 \\
muv-aid466 & 30 & 14999 & GPCR & S1P1 receptor \\
muv-aid852 & 30 & 15000 & protease & FXIIa \\
muv-aid859 & 30 & 15000 & GPCR & M1 receptor \\
tox-NR-Aromatase & 300 & 5521 & other enzyme & Aromatase \\
tox-SR-MMP & 919 & 4891 & miscellaneous & mitochondrial membrane potential \\
\bottomrule
\end{tabular}
\end{table}

\begin{table}[ht]
\centering
\caption{Held-out datasets.}
\label{tab:held-out}
\vskip 0.2in
\begin{tabular}{lSSll}
\toprule
\bfseries Dataset & \bfseries Actives & \bfseries Inactives & \bfseries Target Class & \bfseries Target \\
\midrule
pcba-aid1461 & 2305 & 218561 & GPCR & NPSR \\
pcba-aid2675 & 99 & 279333 & miscellaneous & MBNL1-CUG \\
pcba-aid602233 & 165 & 380904 & other enzyme & PGK \\
pcba-aid624417 & 6388 & 398731 & GPCR & GLP-1 \\
pcba-aid652106 & 496 & 368281 & miscellaneous & alpha-synuclein \\
muv-aid548 & 30 & 15000 & protein kinase & PKA \\
muv-aid832 & 30 & 15000 & protease & Cathepsin G \\
muv-aid846 & 30 & 15000 & protease & FXIa \\
tox-NR-AhR & 768 & 5780 & transcription factor & Aryl hydrocarbon receptor \\
tox-SR-ATAD5 & 264 & 6807 & promoter & ATAD5 \\
\bottomrule
\end{tabular}
\end{table}

\clearpage

\section{Performance metrics}

\begin{figure}[ht]
\centering
\includegraphics[height=0.9\textheight]{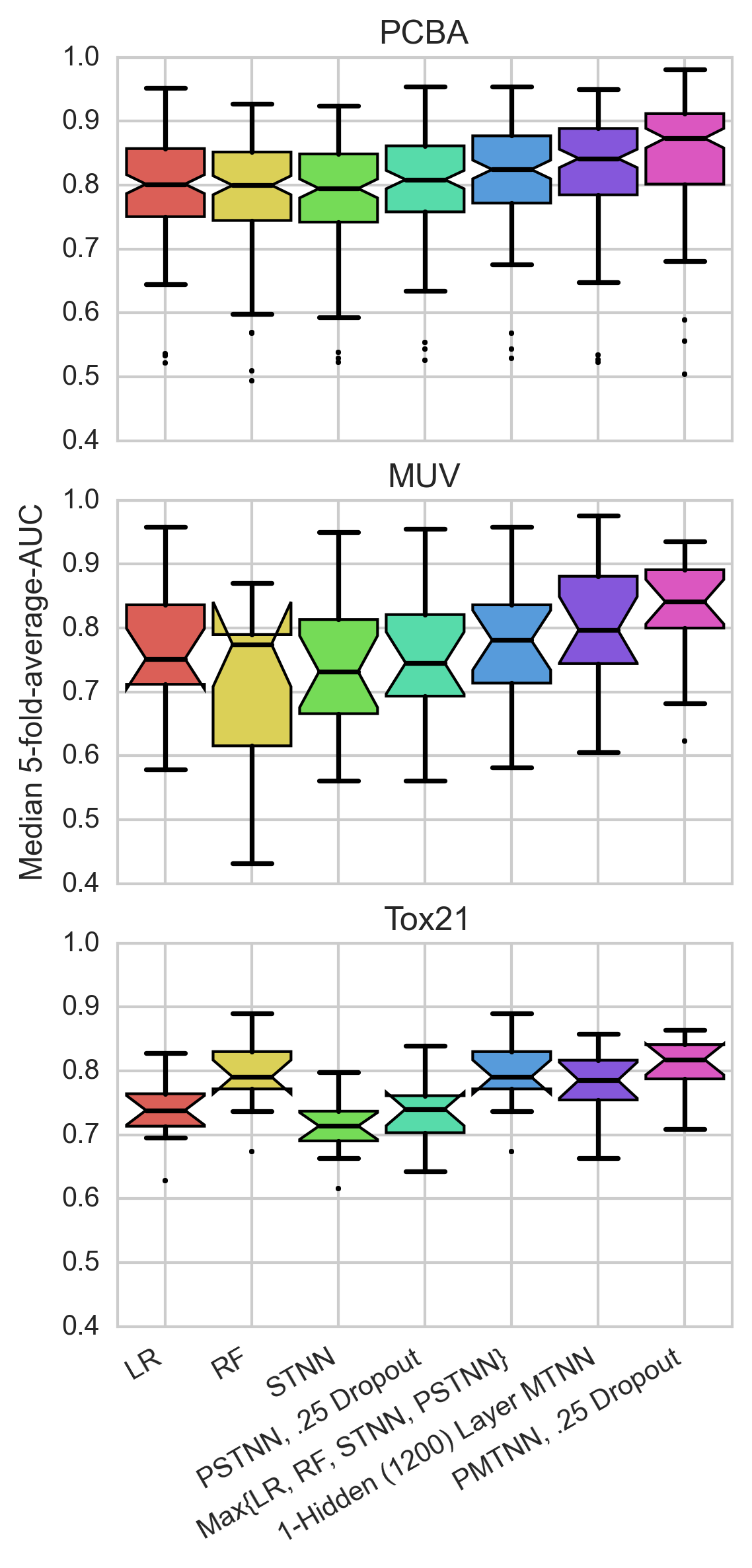}
\caption{Graphical representation of data from Table 2 in the text. Notches
  indicate a confidence interval around the median, computed as $\pm 1.57
  \times \text{IQR}/ \sqrt{N}$ \citep{mcgill1978variations}. Occasionally the
  notch limits go beyond the quartile markers, producing a ``folded down''
  effect on the boxplot. Paired $t$-tests (2-sided) relative to the PMTNN
  across all non-DUD-E datasets gave $p \le \num{1.86e-15}$.}
\label{fig:table2boxplot}
\end{figure}

\begin{table}[ht]
\centering
\caption{Sign test CIs for each group of datasets. Each model is compared
to the Pyramidal $(2000, 100)$ Multitask Neural Net, .25 Dropout model.}
\label{tab:sign-tests}
\vskip 0.2in
\begin{tabular}{ccccc}
\toprule
Model & \makecell{PCBA \\ $(n=128)$} & \makecell{MUV \\ $(n=17)$} & \makecell{Tox21 \\ $(n=12)$} \\
\midrule
Logistic Regression (LR)
& $[.3, .11]$ & $[.13, .53]$ & $[.00, .24]$ \\
Random Forest (RF)
& $[.05, .16]$ & $[.00, .18]$ & $[.14, .61]$ \\
Single-Task Neural Net (STNN)
& $[.02, .10]$ & $[.13, .53]$ & $[.00, .24]$ \\
Pyramidal $(2000, 100)$ STNN, .25 Dropout (PSTNN)
& $[.05, .15]$ & $[.13, .53]$ & $[.00, .24]$ & \\
Max\{LR, RF, STNN, PSTNN\}
& $[.09, .21]$ & $[.13, .53]$ & $[.14, .61]$ & \\
$1$-Hidden $(1200)$ Layer Multitask Neural Net (MTNN)
& $[.05, .15]$ & $[.22, .64]$ & $[.01, .35]$ & \\
\bottomrule
\end{tabular}
\end{table}

\begin{table}[ht]
\centering
\caption{Enrichment scores for all models reported in Table~2. Each value
  is the median across the datasets in a group of the mean $k$-fold
  enrichment values. Enrichment is an alternate measure of model performance
  common in virtual drug screening. We use the ``ROC enrichment'' definition
  from~\cite{jain2008recommendations}, but roughly enrichment is the factor
  better than random that a model's top $X\%$ predictions are.}
\label{tab:enrichment}
\vskip 0.2in
\begin{tabular}{l|cccc|cccc|cccc}
\toprule
Model & \multicolumn{4}{|c|}{PCBA} & \multicolumn{4}{|c|}{MUV} &
 \multicolumn{4}{|c}{Tox21} \\
&
0.5\% & 1\% & 2\% & 5\% &
0.5\% & 1\% & 2\% & 5\% &
0.5\% & 1\% & 2\% & 5\% \\
\midrule
LR
& 19.4 &   16.5 &   12.1 &   7.9
& 20.0 &   23.3 &   15.0 &   8.0
& 23.9 &   18.3 &   10.6 &   6.7
\\
RF
& 40.0 &   27.4 &   17.4 &   9.1
& \textbf{40.0} &   \textbf{26.7} &   \textbf{16.7} &   7.3
& 23.2 &   \textbf{19.5} &   \textbf{13.6} &   7.8
\\
STNN
& 19.0 &   15.6 &   11.8 &   7.7
& 26.7 &   20.0 &   11.7 &   8.0
& 16.2 &   14.4 &   9.8  &   6.1
\\
PSTNN
& 21.8 &   16.9 &   12.4 &   7.9
& 26.7 &   16.7 &   13.3 &   8.0
& 23.8 &   16.1 &   10.0 &   6.7
\\
MTNN
& 33.8 &   23.6 &   16.9 &   9.8
& 26.7 &   16.7 &   \textbf{16.7} &   8.7
& \textbf{24.5} &   18.0 &   11.4 &   6.9
\\
PMTNN
& \textbf{43.8} &   \textbf{29.6} &   \textbf{19.7} &   \textbf{11.2}
& \textbf{40.0} &   23.3 &   \textbf{16.7} &   \textbf{10.0}
& 23.5 &   18.5 &   \textbf{13.7} &   \textbf{8.1}
\\
\bottomrule
\end{tabular}
\end{table}

\clearpage

\section{Training Details}
The multitask networks in Table 2 were trained with learning rate $.0003$
and batch size $128$ for $50$M steps using stochastic gradient descent.
Weights were initialized from a zero-mean Gaussian with standard deviation
$.01$. The bias was initialized at $.5$. We experimented with higher
learning rates, but found that the pyramidal networks sometimes failed to
train (the top hidden layer zeroed itself out). However, this effect
vanished with the lower learning rate.  Most of the models were trained
with 64 simultaneous replicas sharing their gradient updates, but in some
cases we used as many as 256. 

The pyramidal single-task networks were trained with the same settings, but
for $100$K steps. The vanilla single-task networks were trained with
learning rate $.001$ for $100$K steps. The networks used in Figure 3 and
Figure 4 were trained with learning rate $0.003$ for 500 epochs plus a
constant 3 million steps. The constant factor was introduced after we
observed that the smaller multitask networks required more epochs than the
larger networks to stabilize.

The networks in Figure~5 were trained with a Pyramidal (1000, 50) Single
Task architecture (matching the networks in Figure~3). The weights were
initialized with the weights from the networks represented in Figure~3 and
then trained for 100K steps with a learning rate of 0.0003.

As we noted in the main text, the datasets in our collection contained many
more inactive than active compounds. To ensure the actives were given
adequate importance during training, we weighted the actives for each
dataset to have total weight equal to the number of inactives for
that dataset (inactives were given unit weight).

\tablename~\ref{tab:sensitivity} contains the results of our pyramidal
model sensitivity analysis.  Tables \ref{tab:model_descriptions} and
\ref{tab:models} give results for a variety of additional models not
reported in Table 2.

\begin{table}[ht]
\centering
\caption{Pyramid sensitivity analysis. Median 5-fold-average-AUC values are
  given for several variations of the pyramidal architecture. In an attempt
  to avoid the problem of training failures due to the top layer becoming all
  zero early in the training, the learning rate was set to 0.0001 for the
  first 2M steps then to 0.0003 for 28M steps.}
\label{tab:sensitivity}
\vskip 0.2in
\begin{tabular}{l >{$}c<{$} >{$}c<{$} >{$}c<{$} }
\toprule
\textnormal{Model} & \makecell{\text{PCBA} \\ (n=128)} & \makecell{\text{MUV} \\ (n=17)} & \makecell{\text{Tox21} \\ (n=12)} \\
\midrule
Pyramidal $(1000, 50)$ MTNN & .846 & .825 & .799 \\
Pyramidal $(1000, 100)$ MTNN & .845 & .818 & .796 \\
Pyramidal $(1000, 150)$ MTNN & .842 & .812 & .798 \\
Pyramidal $(2000, 50)$ MTNN & .846 & .819 & .794 \\
Pyramidal $(2000, 100)$ MTNN & .846 & .821 & .798 \\
Pyramidal $(2000, 150)$ MTNN & .845 & .839 & .792 \\
Pyramidal $(3000, 50)$ MTNN & .848 & .801 & .796 \\
Pyramidal $(3000, 100)$ MTNN & .844 & .804 & .799 \\
Pyramidal $(3000, 150)$ MTNN & .843 & .810 & .789 \\
\bottomrule
\end{tabular}
\end{table}

\clearpage

\begin{table}[ht]
\centering
\caption{Descriptions for additional models. MTNN: multitask neural net.
  ``Auxiliary heads'' refers to the attachment of independent softmax units
  for each task to hidden layers \cite[see][]{szegedy2014going}. Unless
  otherwise marked, assume 10M training steps.}
\label{tab:model_descriptions}
\vskip 0.2in
\begin{tabular}{>{\bfseries}cl}
\toprule
A & $8$-Hidden $(300)$ Layer MTNN, auxiliary heads attached to hidden layers 3 and 6, 6M steps \\
B & $1$-Hidden $(3000)$ Layer MTNN, 1M steps \\
C & $1$-Hidden $(3000)$ Layer MTNN, 1.5M steps \\
D & Pyramidal $(1800, 100)$, 2 deep, reconnected (original input concatenated to first pyramid output) \\
E & Pyramidal $(1800, 100)$, 3 deep \\
F & $4$-Hidden $(1000)$ Layer MTNN, auxiliary heads attached to hidden layer 2, 4.5M steps \\
G & Pyramidal $(2000, 100)$ MTNN, 10\% connected \\
H & Pyramidal $(2000, 100)$ MTNN, 50\% connected \\
I & Pyramidal $(2000, 100)$ MTNN, $.001$ learning rate \\
J & Pyramidal $(2000, 100)$ MTNN, 50M steps, $.0003$ learning rate \\
K & Pyramidal $(2000, 100)$ MTNN, $.25$ Dropout (first layer only), 50M steps \\
L & Pyramidal $(2000, 100)$ MTNN, $.25$ Dropout, $.001$ learning rate \\
\bottomrule
\end{tabular}
\end{table}

\begin{table}[ht]
\centering
\caption{Median 5-fold-average AUC values for additional models. Sign test
  confidence intervals and paired $t$-test (2-sided) $p$-values are relative
  to the PMTNN from Table 2 and were calculated across all non-DUD-E
  datasets.}
\label{tab:models}
\vskip 0.2in
\begin{tabular}{ >{\bfseries}c >{$}c<{$} >{$}c<{$} >{$}c<{$} >{$}c<{$} >{$}c<{$} }
\toprule
\textnormal{Model} & \makecell{\text{PCBA} \\ (n=128)} & \makecell{\text{MUV} \\ (n=17)} & \makecell{\text{Tox21} \\ (n=12)} & \text{Sign Test CI} & \text{Paired $t$-Test} \\
\midrule
A & .836 & .793 & .786 & [.01, .06] & \num{9.37e-43} \\
B & .835 & .855 & .769 & [.11, .22]  & \num{1.17e-17} \\
C & .837 & .851 & .765 & [.12, .24] & \num{2.60e-16} \\
D & .842 & .842 & .816 & [.08, .18] & \num{1.89e-21} \\
E & .842 & .808 & .789 & [.02, .08] & \num{9.25e-43} \\
F & .858 & .836 & .810 & [.10, .22] & \num{4.85e-13} \\
G & .831 & .795 & .774 & [.03, .11] & \num{1.15e-31} \\
H & .856 & .827 & .796 & [.04, .13] & \num{5.34e-21} \\
I & .860 & .862 & .824 & [.07, .17] & \num{6.23e-14} \\
J & .830 & .810 & .801 & [.05, .14] & \num{9.25e-25} \\
K & .859 & .843 & .803 & [.24, .38] & \num{3.25e-9} \\
L & .872 & .837 & .802 & [.35, .50] & \num{2.74e-2} \\
\bottomrule
\end{tabular}
\end{table}

\putbib
\end{bibunit}
\end{document}